# Joint discovery of governing partial differential equations from multi-source datasets by competitive optimization

Hao Xu[1,2], Siyu Lou[1,3], Yuntian Chen[1,4,*], and Dongxiao Zhang[1,*]

[1] Zhejiang Key Laboratory of Industrial Intelligence and Digital Twin, Eastern Institute of Technology, Ningbo, Zhejiang 315200, China

[2] Department of Electrical Engineering, Tsinghua University, Beijing 100084, P. R. China.

[3] Shanghai Jiao Tong University, School of Computer Science, Shanghai, China

[4] Ningbo Institute of Digital Twin, Eastern Institute of Technology, Ningbo, Zhejiang 315200, P. R. China

[*] Corresponding authors.

Email address: ychen@eitech.edu.cn (Y. Chen); dzhang@eitech.edu.cn (D. Zhang).

**Abstract**

Discovering governing equations directly from observational data is a key step towards interpretable scientific machine learning. Current data-driven approaches typically operate on a single dataset, inherently limiting their performance when faced with restricted observations. In practice, multiple datasets are often available for the same physical system, distinguished only by distinct initial conditions or boundary configurations. Here, we present a competitive optimization framework designed to discover shared partial differential equations (PDEs) from multi-source datasets, termed MCO-PDE. The framework first trains independent neural surrogates for each data source, and then employs a soft-competitive weighting mechanism to dynamically assess dataset credibility and aggregate a consensus global coefficient. Integrated with a genetic algorithm for structural search, this approach simultaneously identifies the functional forms and parameters of the governing laws. We demonstrate that fusing as few as 50 observations per dataset across seven cases recovers canonical equations with high accuracy. The framework inherently handles two- and three-dimensional domains characterized by irregular boundaries and heterogeneous coefficients, and successfully extracts physically meaningful laws from real-world wave-tank experiments. Overall, this work establishes a promising route for automated scientific discovery via heterogeneous data fusion.

## Introduction

Modern science rarely encounters a phenomenon through a single window. The same process is typically observed under distinct conditions, each yielding a partial view shaped by its measurement setting. For instance, replicated experiments conducted across independent chemistry laboratories yield varying readings under identical governing kinetics; similarly, discrete contaminant monitoring stations record disparate release histories while sampling the exact same migration process. Reconciling these views into coherent knowledge is the central promise of data fusion, which is in increasing demand as observational modalities multiply. This gives rise to what we term multi-source knowledge discovery, in which heterogeneous datasets collected from distinct sources are integrated to uncover the common underlying law.

For the discovery of governing partial differential equations (PDEs) from a single dataset, substantial progress has been made[1], evolving from sparse regression on fixed libraries to increasingly open and generative searches over equation space. Early approaches, represented by SINDy[2] and PDE-FIND[3], recover PDEs by selecting a sparse set of active terms from a predefined library, and later variants such as WSINDy[4] and WeakIdent[5] improve robustness by using weak or integral formulations to reduce sensitivity to noise and numerical differentiation errors[6,7]. The introduction of genetic algorithms further relaxes the closed-library setting to a semi-open space of candidate structures[8,9]. Meanwhile, the incorporation of flexible symbolic-regression methods moves beyond predefined combinations and enable freer exploration of equation forms[10–12]. In parallel, neural-network surrogates, as used in PDE-Net [13,14], DeepMod [15] and DL-PDE[16], made it possible to infer latent fields and compute derivatives by automatic differentiation, thereby reducing reliance on finite-difference estimates. Building on this trend, PINN-based frameworks coupled knowledge embedding and equation discovery through alternating optimization [17–19], improving robustness under sparse and noisy observations. Most recently, generative models[20] and large language models[21,22] have begun to guide equation search with learned priors. These methods have enabled the rediscovery of canonical equations and the identification of previously unknown physical laws[23,24].

However, all of the above methods are designed for and evaluated on single-source data. When the number of observations per experiment is limited, single-dataset methods exhibit markedly degraded performance and diminished stability in equation discovery. Faced with multiple datasets, a naïve strategy would be to apply these individual methods independently to each source and subsequently reconcile the results, such as through majority voting. However, independent identification on sparse datasets frequently yields disparate and often erroneous equations, rendering a clear consensus elusive. While recent attempts have integrated group sparsity into frameworks like DeepMod to handle multiple datasets[25], they primarily rely on shared term selection rather than deeper cross-dataset fusion, and thus remain constrained by the need for relatively dense observations. This situation is analogous to the Chinese parable of the blind men and the elephant (Fig. 1a). Each observer, restricted to a localized perspective, reconstructs a different global model from limited data. For instance, one blind man examining the tail misinterprets the system as a rope ($u_{xxx}$); while another, touching the leg, concludes it is a tree ($uu_x$). Only by aggregating these complementary, partial views can the true form of the elephant (i.e., the complete underlying physics) be fully revealed. However, traditional data-fusion methodologies have been developed primarily for classification and state estimation, relying on frameworks such as Bayesian inference and Dempster–Shafer theory[26]. While modern machine learning-based techniques, such as multi-source domain adaptation[27]exploit heterogeneous sources to improve predictive accuracy[28], their output remains a black-box predictor or a latent representation rather than a closed-form description of the underlying process. Consequently, extending equation discovery to multi-source data introduces three fundamental obstacles: (i) the equation structure is unknown and must be searched over a combinatorically large space; (ii) derivative estimation from limited data is inherently ill-conditioned; and (iii) datasets often exhibit heterogeneous quality, meaning an unweighted treatment leads to low-quality sources that will corrupt the global estimate.

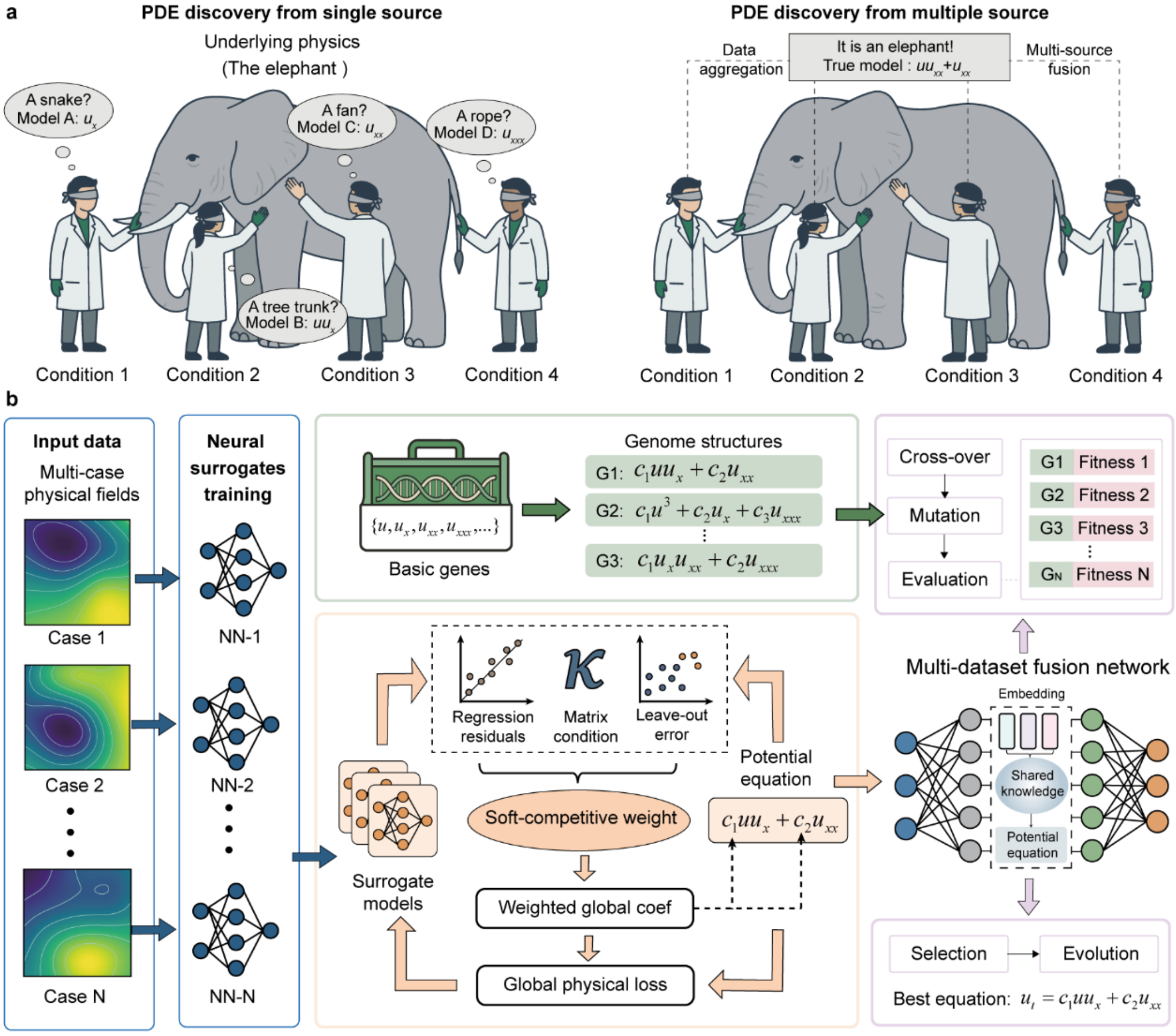

**Fig. 1. Schematic of multi-source PDE discovery. (a),** Analogy of blind observations of an elephant to illustrate the difference between single-source and multi-source PDE discovery. In the single-source setting, each condition provides only a partial view of the underlying physics and may lead to different candidate models. Conversely, integrating complementary information across multiple diverse sources breaks through these localized blind spots, enabling the robust recovery of the shared governing equation. (**b**)**,** Overview of the proposed competitive optimization framework for multi-source PDE discovery (MCO-PDE). First, multi-case physical fields are fitted via independent neural surrogates. Candidate functional terms are then encoded as fundamental genes and assembled into genomic structures, which are evaluated using the trained surrogates. Next, a soft-competitive weighting strategy dynamically estimates global coefficients and computes a unified physical loss for each candidate equation. Finally, the optimal PDE architecture is distilled through an evolutionary search procedure.

To address these challenges, we propose a competitive optimization framework for multi-source PDE discovery, which is termed as MCO-PDE (Fig. 1b). The core idea is to train an independent neural surrogate for each dataset, compute derivatives via automatic differentiation, and then jointly evaluate candidate equation structures across all datasets using a soft-competitive weighting mechanism. Rather than treating every dataset equally, MCO-PDE assigns each dataset a credibility score based on three diagnostics, including the regression residual of the PDE, the condition number of the design matrix, and a leave-out generalization error. These scores are converted into softmax weights that determine each dataset's contribution to a global coefficient estimate; the global coefficients in turn drive a unified PDE loss that regularizes all surrogates simultaneously. An exponential moving average (EMA) of the global coefficients prevents oscillation and ensures stable convergence. The equation-structure search is performed by a genetic algorithm that generates and evolves candidate genome representations of PDEs based on the competitive optimization outcomes. The MCO-PDE framework is validated on a broad suite of canonical PDEs ranging from one to three spatial dimensions, which proves that fusing multiple sources markedly reduces the observations required from any single case. Beyond synthetic benchmarks, the framework discovers physically interpretable equations from real-world multi-source wave-tank experiments. These findings indicate that the proposed framework is promising for scientific knowledge discovery, particularly in settings where observations are constrained but their origins are rich.

## Results

### Discovery of the canonical PDEs from multi-source dataset under sparse and noisy data

We first demonstrate the procedure of the MCO-PDE framework by the discovery of the Burgers' equation, which is written as:

$$u_t = -uu_x + 0.1u_{xx} \tag{1}$$

Seven datasets are generated from different initial conditions on a grid of $N_x = 256$ spatial points and $N_t = 201$ time steps (Fig. 2a). Each dataset contains 51,456 total observations, from which we randomly sample a small number of data points per case for the PDE discovery. A detailed description of the utilized datasets is provided in Supplementary Information S1.1. With only 50 randomly sampled observations per dataset, MCO-PDE successfully identifies the correct equation structure $u_t=c_1uu_x+c_2u_{xx}$ and recovers coefficients $c_1 = -0.957 \pm 0.011$ and $c_2 = 0.102 \pm 0.003$, in close agreement with the true values of –1.0 and 0.1. Currently, no existing single-dataset method achieves this accuracy at such limited data volume. Notably, MCO-PDE simultaneously identifies the equation structure, estimates the associated coefficients, and quantifies their uncertainty. Such uncertainty estimation has been difficult to achieve within single-source discovery frameworks, it is nevertheless crucial for quantifying the confidence and reliability of the identified equations. From Fig. 2b, the genetic-algorithm loss is steadily optimized and converges within 8 generations. Because MCO-PDE does not impose sparsity constraints during optimization, the genetic algorithm typically returns a range of candidate equation structures, some of which may contain redundant terms. We therefore introduce a physics-informed information criterion (PIC) to evaluate all substructures of these candidates in terms of both physical fidelity and mathematical parsimony by the loss and mean coefficient of variation[29], thereby identifying the most appropriate final equation. It can be seen that the correctly discovered PDE structure has the lowest PIC value (Fig. 2b).

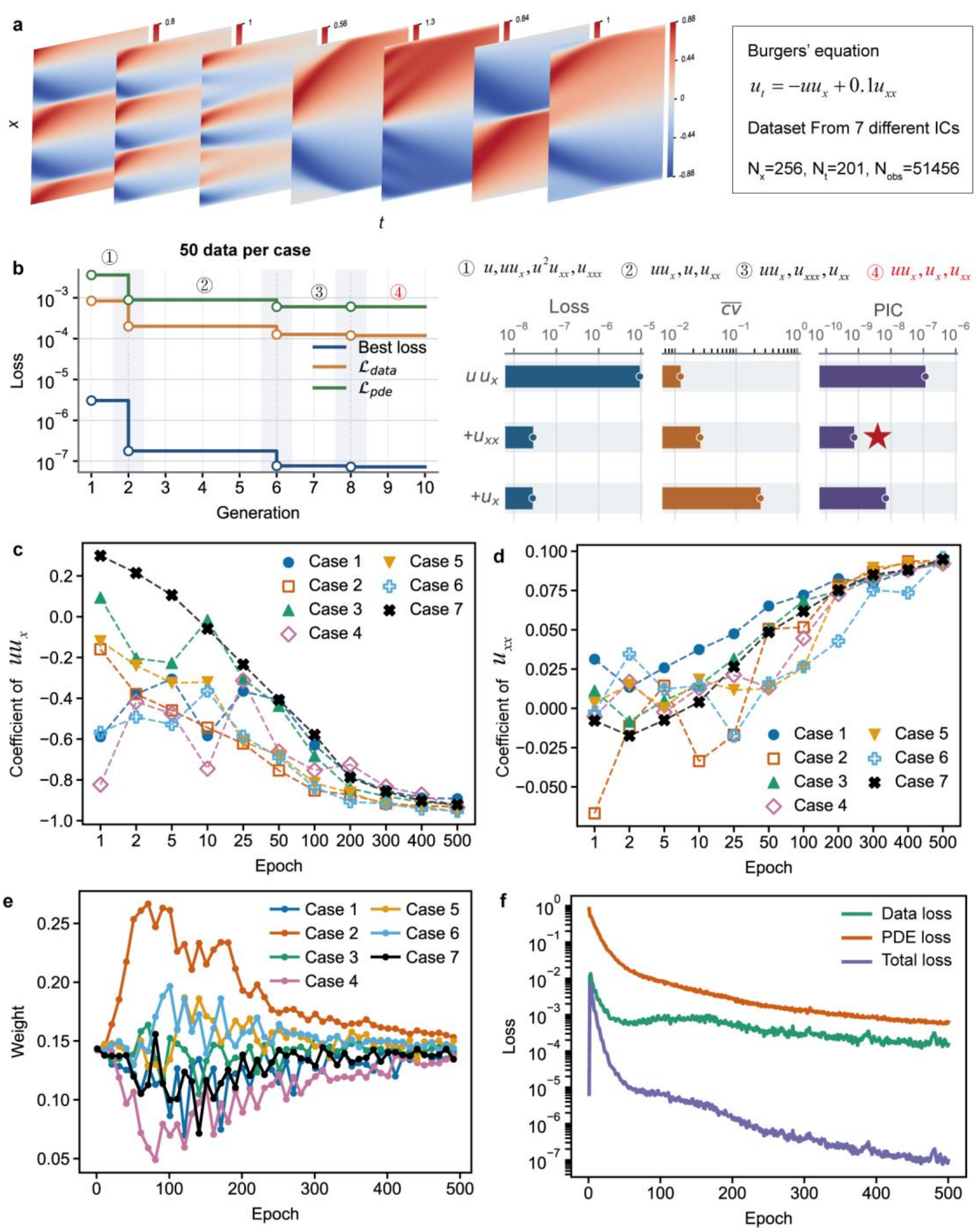

**Fig. 2. Discovery of the Burgers' equation from multi-source dataset. (a),** Seven datasets generated from the Burgers' equation under different initial conditions. **(b),** Evolutionary search and model selection, including best loss, data loss and PDE loss across generations (left) and comparison of candidate substructures using the physics-informed information criterion (right). **(c,d),** Convergence of the case-specific coefficients associated with $uu_x$ and $u_{xx}$, respectively, during competitive optimization across the seven datasets. **(e),** Evolution of the dataset weights during the competitive optimization. **(f),** Training dynamics of the global objective, including data loss, PDE loss and overall loss.

The competitive weighting mechanism plays a crucial role in the discovery success. Fig. 2c and 2d show the evolution of the per-dataset coefficients for $uu_x$ and $u_{xx}$ over the training epochs. Initially, the seven datasets produce widely scattered coefficient estimates, reflecting the high variability inherent in sparse data. As training progresses, the soft-competitive weights (Fig. 2e) dynamically adjust. Datasets featuring well-converged surrogates and highly consistent coefficient estimates are automatically prioritized with higher weights. In contrast, under-fitted sources are suppressed, minimizing their influence on the global estimate until their reliability improves. By epoch 300, the coefficient estimates across all seven individual datasets converge toward a unified consensus, accompanied by the stabilization of their respective weights. Concurrently, the data loss, PDE residual, and overall loss score exhibit a synchronized, monotonic decline (Fig. 2f), confirming the robust convergence of the entire optimization process. In the Supplementary Information, we further compare the training dynamics for the correct structure $\{uu_x, u_{xx}\}$ against a wrong structure $\{uu_{xx}, u_x\}$ (Fig. S1). Conversely, the latter setup yields divergent per-dataset coefficients and oscillating weights, with the final loss plunging by three orders of magnitude. This distinct failure profile provides an unambiguous signal for model selection.

For a greater number of observations per dataset, the identified coefficients will be more accurate with a smaller uncertainty. With 1000 data per case, the discovered equation is $u_t$ = (–0.980 ± 0.011) $uu_x$ + (0.106 ± 0.001) $u_{xx}$. When confronted with sparse and noisy data, MCO-PDE also performs well. With 100 observations per case and 5% Gaussian noise, it recovers $u_t$ = (–0.926 ± 0.027) $uu_x$ + (0.106 ± 0.002) $u_{xx}$. With 200 observations per case and 10% noise, the coefficients are (–0.835 ± 0.034) and (0.118 ± 0.005). These results show that uncertainty increases under smaller sample sizes and higher noise levels, yet MCO-PDE still recovers the correct governing equation.

To assess the generality and robustness of MCO-PDE on PDEs with higher-order derivatives, we then test it on the Korteweg–de Vries (KdV) equation, which is written as:

$$u_t = -uu_x - 0.0025u_{xxx}, \tag{2}$$

For the KdV equation, seven datasets are generated from different initial conditions on a grid of 512×201 points (Fig. S2). With only 50 observations per case, the correct PDE can be identified from seven datasets:

$$u_t = -(0.867 \pm 0.008)\, uu_x - (0.0023 \pm 4.32 \times 10^{-5})\, u_{xxx}, \tag{3}$$

Table S1 summarizes the performance of MCO-PDE across different numbers of datasets and data volumes per dataset. These results indicate that increasing the number of datasets markedly improves the accuracy of PDE discovery. Using a single dataset yields an incorrect equation $u_t = -0.702u+0.083u_x$, but with four or more datasets, the correct structure can be identified. Meanwhile, increasing the data volume to 1000 points per case yields near-exact discovery, $u_t = -(0.950 \pm 0.008)\ uu_x - (0.0025 \pm 1.19 \times 10^{-5})\ u_{xxx}$. This progressive improvement with increasing dataset count provides strong empirical evidence for the value of multi-source fusion.

Then, we consider the discovery of Allen–Cahn equation, which presents a more complex form with a cubic reaction term:

$$u_t = u - u^3 + 0.003u_{xx}, \tag{4}$$

Seven datasets are generated with different initial conditions and spatio-temporal domains (Fig. S3). With 200 observations per case, MCO-PDE discovers:

$$u_t = (0.832 \pm 0.113)\ u - (0.769 \pm 0.153)\ u^3 + (0.0028 \pm 1.49 \times 10^{-4})\ u_{xx}, \tag{5}$$

It correctly identifies the three-term structure despite the presence of a very small diffusion coefficient that is easily masked by noise. With 1000 observations, the coefficients will be identified more accurately with $(0.957 \pm 0.010)\ u$, $(-0.941 \pm 0.012)\ u^3$, and $(0.0029 \pm 3.40 \times 10^{-5})\ u_{xx}$.

Finally, the discovery of Klein–Gordon (KG) equation is investigated, which has a second-order temporal derivative on the left-hand side.

$$u_{tt} = -5u+0.5u_{xx}, \tag{6}$$

Seven datasets are generated from different initial conditions (Fig. S4). With 100 observations per case, the correct PDE can be identified, which is written as:

$$u_{tt} = -(4.39 \pm 0.052)\ u+(0.475 \pm 0.002)\ u_{xx}, \tag{7}$$

As the data volume increases to 1000 per case, the accuracy of identified coefficients will be improved to $-(4.84 \pm 0.062)\ u$ and $(0.482 \pm 8.68 \times 10^{-4})\ u_{xx}$. These results demonstrate that MCO-PDE enables reliable discovery of canonical PDEs from sparse and noisy multi-source datasets.

## Comparison with existing methods

To contextualize the performance gains enabled by multi-source fusion, we compare MCO-PDE against seven established single-dataset PDE discovery methods: PDE-FIND[3], WSINDy[4], WeakIdent[5], PDE-READ[30], GGA[17], DISCOVER[10], and ABL-PDE[31]. All methods are applied to the same seven Burgers' equation datasets with 50, 100, and 1000 observations per case. For a fair comparison, spatial and temporal derivatives for all baselines were obtained by automatic differentiation of surrogate models trained with the corresponding number of observations in each case. Derivatives were evaluated on a metadata grid of 10,000 uniformly sampled points, constructed from 100 points in $x$ and 100 points in $t$ over the computational domain. PDE-READ was treated separately, because it includes its own neural-network fitting procedure and was therefore applied directly to the original observations. All conventional methods were run using their original open-source implementations, and hyperparameters were kept at the default settings used in their Burgers equation examples. To extend these conventional single-dataset PDE discovery methods to the multi-source setting, we adopted a voting-based aggregation strategy. Specifically, each dataset was analyzed independently, and the equation structure that appeared most frequently across the discovery process was taken as the final PDE. The results are provided in Table 1.

When 1,000 observations are available for each case, most existing methods perform well, indicating that PDE discovery becomes easier once data are sufficient. In this regime, the identified equations are also relatively consistent across datasets. For example, WSINDy, GGA and DISCOVER recover the correct Burgers equation in all seven cases, while PDE-FIND, WeakIdent, ABL-PDE and PDE-READ succeed in most cases. Even when some methods fail on a subset of datasets, a simple voting strategy can still recover the correct final equation because the dominant pattern across cases remains stable. The situation changes markedly when the number of observations is reduced to 100 per case and 50 per case. In the condition of sparse data, none of the conventional methods discovers the correct equation across all cases. A likely reason is that the reduced sample size substantially degrades derivative estimation, thereby increasing both the difficulty and the uncertainty of equation discovery. Consequently, the equations identified across different scenarios become highly inconsistent, as exemplified by methods like WSINDy and GGA, rendering majority voting unreliable under these conditions. For ABL-PDE and DISCOVER, three of the seven individual cases converge onto the exact same incorrect structure. This failure underscores a fundamental vulnerability of post hoc aggregation: it inherently privileges statistical frequency over physical validity, leaving it incapable of exploiting the complementary information distributed across datasets

**Table 1. Comparison with conventional PDE discovery methods on the discovery of Burgers' equation from multiple-source dataset.** The number in parentheses indicates the occurrence count of the reported structure among the seven independent discoveries.

| Methods | Discovered PDE (50 data per case) | Discovered PDE (100 data per case) | Discovered PDE (1000 data per case) |
|---|---|---|---|
| PDE-FIND | $u_t=C_1u_x$ (1/7) | $u_t=C_1uu_x+C_2u_x$ (2/7) | $u_t=C_1uu_x+C_2u_{xx}$ (5/7) |
| WSINDy | $u_t=C_1uu_x+C_2u_x$ (1/7) | $u_t=C_1uu_x+C_2u_x$ (1/7) | $u_t=C_1uu_x+C_2u_{xx}$ (7/7) |
| WeakIdent | $u_t=C_1u_x$ (2/7) | $u_t=C_1u$ (2/7) | $u_t=C_1uu_x+C_2u_{xx}$ (6/7) |
| PDE-READ | $u_t=C_1uu_x$ (3/7) | $u_t=C_1u$ (3/7) | $u_t=C_1uu_x+C_2u_{xx}$ (2/7) |
| GGA | $u_t=C_1u^2_{xx}+C_2u_{xxx}$ (1/7) | $u_t=C_1u_{xx}+C_2u_{xxxx}$ (1/7) | $u_t=C_1uu_x+C_2u_{xx}$ (7/7) |
| DISCOVER | $u_t=0$ (2/7) | $u_t=C_1u_x$ (3/7) | $u_t=C_1uu_x+C_2u_{xx}$ (7/7) |
| ABL-PDE | $u_t=C_1u_x$ (4/7) | $u_t=C_1u_x$ (3/7) | $u_t=C_1uu_x+C_2u_{xx}$ (5/7) |
| Ours (MCO-PDE) | $u_t=C_1uu_x+C_2u_{xx}$ | $u_t=C_1uu_x+C_2u_{xx}$ | $u_t=C_1uu_x+C_2u_{xx}$ |

A further practical difficulty is that conventional methods typically require case-dependent hyperparameter adjustment, for example in the strength of sparsity regularization. Such tuning becomes particularly problematic in the multi-source setting, where different datasets may favour different parameter choices. By contrast, our competitive optimization neural network performs joint discovery directly from all sources, without case-by-case sparsity-threshold retuning. Through adaptive weighting, it automatically balances the contribution of each dataset and uses all available information in a unified optimization process. This enables accurate discovery of the governing equation even in the low-data regime, and provides a clear advantage over voting-based strategies that simply prioritize the most frequent candidate.

## Discovery of PDEs from higher-dimensional data and irregular boundaries

We next examine the performance of MCO-PDE on higher-dimensional problems. First, we consider a two-dimensional contaminant transport system governed by:

$$u_t = -0.2u_x - 0.15u_y + 0.05u_{xx} + 0.05u_{yy}, \tag{8}$$

Within a rectangular domain defined by $x \in [0,1.5]$ and $y \in [0,1]$, we simulate the spatiotemporal evolution of the contaminant field under different source locations and source geometries, as shown in Fig. 3a. This setting replicates physical scenarios where distinct discharge events yield heterogeneous observations that nonetheless manifest the same underlying governing law. The objective is to reconstruct this shared equation directly from the aggregated multi-source data. Here, the temporal span $t \in [0,1]$, with $n_x$=100, $n_y$=101, and $n_t$=146, so that each case contains nearly 1.5 million spatiotemporal data points. From each case, we randomly sample only 3,000 observations, corresponding to about 0.2% of the full dataset. These observations are therefore extremely sparse over the full space-time domain. As shown in Fig. 3a, despite this severe sparsity, MCO-PDE successfully recovers the correct equation structure by integrating information from all seven cases, while maintaining high coefficient accuracy. This result shows that the underlying PDE can still be identified from remarkably limited data even in two spatial dimensions. Notably, conventional single-source PDE discovery methods typically require far denser sampling in high-dimensional settings to reliably identify the correct terms[29]. When the number of observations is increased to 10,000 per case, MCO-PDE remains robust under Gaussian noise levels of up to 10%, still discovering the correct equation with accurate coefficients:

$$u_t = -(0.178 \pm 0.012)u_x - (0.117 \pm 0.008)u_y + (0.040 \pm 7.8\times10^{-4})u_{xx} + (0.039 \pm 5.2\times10^{-4})u_{yy}, \tag{9}$$

Next, the discovery of PDEs on irregular computational regions is investigated. Many conventional frameworks are predominantly designed for regular rectangular grids, where numerical differentiation, weak-form integration and spectral discretization are straightforward to implement[4,32]. In contrast, physical observations are frequently restricted to non-convex domains and irregular boundaries. Such geometric variations, coupled with non-uniform data distributions across different cases, introduce spatial heterogeneity that confounds conventional multi-source PDE discovery. To evaluate MCO-PDE under such conditions, we consider the two-dimensional diffusion equation:

$$u_t = 0.2u_{xx} + 0.2u_{yy}. \tag{10}$$

The datasets are solved on seven distinct irregular computational regions, such as circular, annular, L-shaped and star-shaped domains, each with a different initial condition, as shown in Fig. 3b. In this setting, the definitions of $x$, $y$ and $t$ vary across cases through the domain geometry and admissible sampling region, so that the resulting datasets differ substantially in both shape and size. The total number of spatiotemporal points ranges from $10^5$ to $10^6$ per case. Despite this complexity, MCO-PDE remains effective. With only 500 observations per case, it recovers:

$$u_t = (0.153 \pm 0.002)\, u_{xx} + (0.161 \pm 0.002)\, u_{yy}. \tag{11}$$

It correctly identifying the two-dimensional diffusion structure across all irregular domains. This performance is notable given that conventional methods typically require around twentyfold more data in comparable irregular-domain settings[20]. Increasing the number of observations to 5,000 per case further improves the coefficient estimates to $(0.158 \pm 0.002)\, u_{xx}$ and $(0.180 \pm 0.002)\, u_{yy}$, respectively.

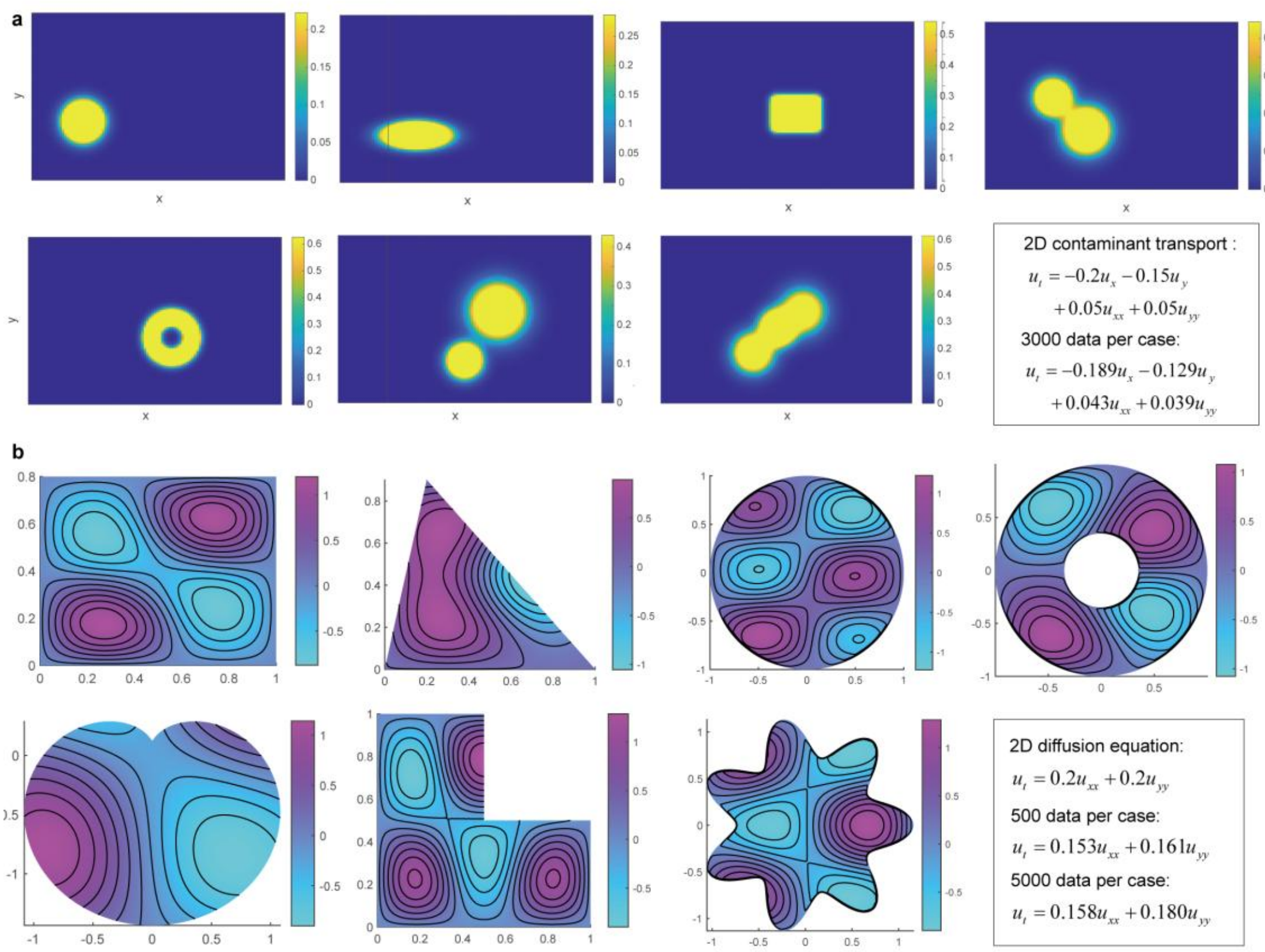

**Fig. 3. Discovery of PDEs from multiple datasets in higher-dimensional and irregular settings. (a)**, Seven two-dimensional contaminant transport cases with different source locations and geometries, and the discovered PDE under 3,000 data per case. **(b)**, Seven two-dimensional diffusion cases defined on irregular computational regions with different shapes and initial conditions, and the discovered PDE under 500 data and 5,000 data per case.

We finally examine whether MCO-PDE remains effective in three spatial dimensions. To this end, we consider the three-dimensional heat equation, which is written as:

$$u_t = 0.01u_{xx} + 0.01u_{yy} + 0.01u_{zz}, \tag{12}$$

The system is solved on a cubic domain with $x$, $y$, $z \in [0,1]$ and $t \in [0,2)$, using grids of 41×41×41 in space and 23 points in time, yielding 1,585,183 spatiotemporal data points per case. Seven datasets are generated under different initial conditions (Fig. S5). Despite the increase in dimensionality, MCO-PDE remains effective under limited observation data. With only 1,000 data points per case, it successfully recovers the correct equation written as:

$$u_t=(0.0076\pm1.8\times10^{-4})\,u_{xx}+(0.0094\pm1.6\times10^{-4})\,u_{yy}+(0.0085\pm1.8\times10^{-4})\,u_{zz}, \quad (13)$$

The identified coefficients remain close to the ground truth despite the extreme sparsity of the sampled observations relative to the full four-dimensional field. With 10,000 observations per case, MCO-PDE is robust to 10% Gaussian noise (Fig. S5). Even in three dimensions, the multi-source approach requires much fewer observations per case than single-case methods, confirming that the fusion mechanism effectively compensates for the curse of dimensionality in data-driven PDE discovery.

**Discovery of heterogeneous PDE from multiple-source datasets**

We next assess the proposed MCO-PDE framework on a more challenging problem, discovery of PDEs in heterogeneous media. We consider two-dimensional transient saturated groundwater flow in porous media, governed by:

$$u_t = 0.001(Ku_x)_x + 0.001(Ku_y)_y, \quad (14)$$

In this system, the governing equation contains a spatially varying heterogeneous field $K$, which introduces additional variables and greatly enlarges the space of possible term combinations, especially in higher-dimensional settings. Meanwhile, the heterogeneous field $K$ appears inside the differential operator, so the governing structure cannot be identified directly from simple candidate terms. This difficulty is further amplified in the multi-source setting, because each case is associated with a different realization of the heterogeneous field, leading to different flow patterns even though the underlying governing law is the same. Here, the computational domain is a square of size 1020×1020, discretized into 51×51 grid blocks, with a total simulation time of 10 and a time step of 0.2, yielding 50 time steps. The hydraulic conductivity field is assumed to be log-normal, with mean $\langle \ln K \rangle = 0$, variance $\sigma^2_{\ln K} = 1$, and correlation length 408. To represent the heterogeneous field, $K$ is parameterized using a Karhunen–Loève expansion with 20 retained terms. Seven cases are generated using different realizations of the conductivity field while keeping the same initial condition, thereby producing a heterogeneous multi-source benchmark (Fig. 4). Because the system involves two variables, the candidate library must account for compound terms involving both the state variable $u$ and the random field $K$. Accordingly, the expanded basic genes is $\{u, u_x, u_{xx}, u_{xxx}, K, K_x, K_{xx}\}$. To control the enlarged search space, we further impose a symmetry constraint during equation generation, so that the discovered structure remains symmetric in $x$ and $y$, consistent with the underlying physics.

With only 500 observations per case, MCO-PDE correctly identifies the expanded form of the heterogeneous governing equation. The discovered equation is written as:

$$u_t = 8.85\times10^{-4}\,Ku_{xx} + 8.64\times10^{-4}\,Ku_{yy} + 7.57\times10^{-4}\,K_xu_x + 5.07\times10^{-4}\,K_yu_y, \quad (15)$$

Although the estimated coefficients of $K_yu_y$ exhibit relatively large discrepancy, this level of accuracy is still notable given the complexity of the process and the extreme sparsity of the data, with only 500 observations per case, corresponding to 0.38% of the full dataset. As the number of observations increases to 1,000 and 5,000 per case, the coefficient estimates become progressively more accurate (Fig. 4). At 5,000 observations per case, all four coefficients fall within 4%–12% of the true value 0.001, with standard deviations on the order of $10^{-5}$. These results show that MCO-PDE can recover parametric PDEs from moderately sized multi-source datasets, even when each source is governed by a different heterogeneous field realization. Conventional PDE discovery methods are rarely applicable to this kind of problem, since they are difficult to jointly resolve hidden coefficient-field effects, operator expansions and cross-source heterogeneity within a unified discovery process.

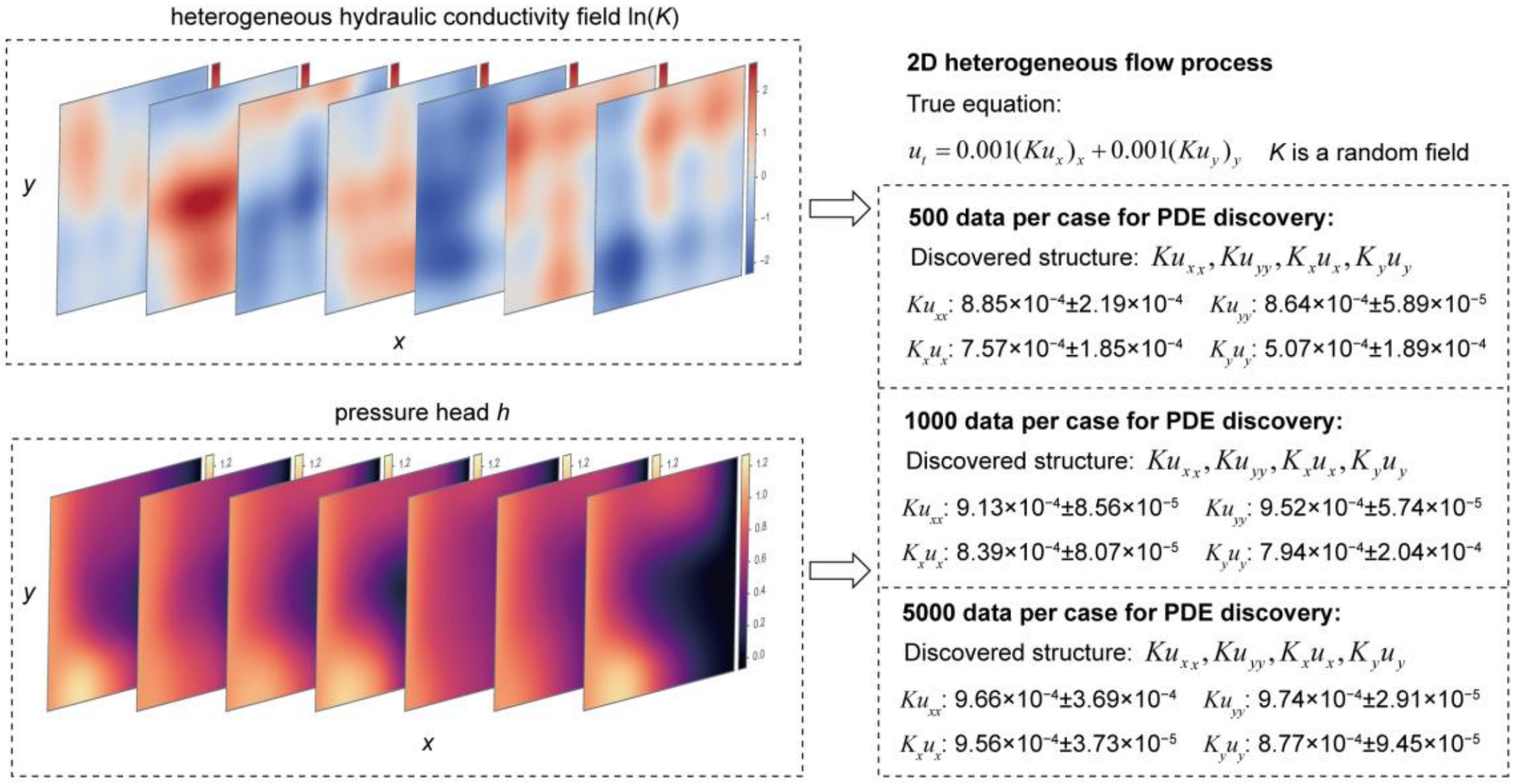

**Fig. 4. Discovery of the governing equation of heterogeneous groundwater flow process from multi-source data.** Seven cases of two-dimensional heterogeneous flow generated from different realizations of the hydraulic conductivity field *K*. The top row shows the spatial heterogeneity of the hydraulic conductivity field, and the bottom row shows the corresponding pressure-head responses. The right panel summarizes the discovered equation structures and coefficient estimates obtained from different numbers of observations per case, including 500 data, 1,000 data and 5,000 data.

## Discovery of governing equations from real-world experimental data

Finally, we tested the ability of the proposed MCO-PDE framework to discover interpretable governing equations directly from laboratory measurements. The experiments were conducted in a 27.2-m wave tank equipped with a wave paddle and three CCD cameras[33,34] (Fig. 5a). Because the supporting columns partially blocked the optical path, the observable free-surface data were not continuous in space but split into three visible sub-intervals within the camera-covered region from 8.0 to 12.5 m. To construct a multi-source setting under controlled conditions, we performed four independent runs with different paddle amplitudes, *A*=80, 90, 100 and 105, while keeping the remaining experimental conditions unchanged. These four experimental trials were designated as separate cases sharing identical underlying wave dynamics but differing in their excitation strengths. The heatmaps of the datasets are provided in Fig. S6.

A practical limitation of this dataset is that the raw observations are spatially incomplete for direct derivative estimation. To address this, we used only 40% of the available spatio-temporal samples in each case to train surrogate models, and then evaluated the required derivatives over the full observed $x$-domain from the generated meta-data. For PDE discovery, we further non-dimensionalized the variables as:

$$x^* = \frac{x-x_0}{\lambda}, \quad t^* = \frac{t}{T_p}, \quad \eta^* = \frac{\eta}{\lambda}, \tag{16}$$

where $x_0$ is the starting coordinate of the observed region, $\lambda$ is the initial wavelength, $T_p$ is the initial peak period, and $\eta$ is the surface elevation. For each case, we then estimated the onset of the near-breaking regime from the spatio-temporal wave field itself. Specifically, we tracked the evolution of the maximum spatial slope,

$$S(t) = \max_x \left|\frac{\partial \eta}{\partial x}\right|, \tag{17}$$

which quantifies the strongest instantaneous front steepness in the profile. When $S(t)$ rose markedly above its sliding average over a preceding time window, that time was identified as the onset of near breaking. When no clear threshold jump was observed, we used the time at which the growth rate of $S(t)$ attained its maximum as an auxiliary criterion. This data-driven partition allowed each run to be divided into two physically distinct stages: a pre-steepening stage, in which the wave mainly propagates with only mild nonlinear deformation, and a pre-breaking stage, in which nonlinear steepening becomes dominant (Fig. 5b).

For the pre-steepening stage, the jointly discovered PDE from all four cases is written as:

$$\eta_t^* + 0.72\eta_x^* + 2.36 \times 10^{-4}\eta_{xxx}^* = 0, \tag{18}$$

The competitive optimization and model selection process is illustrated in Fig. 5c and Fig. S7a. This equation is physically consistent with a weakly nonlinear long-wave regime. The dominant first-derivative term represents bulk wave propagation, whereas the much smaller third-derivative term provides a dispersive correction. The absence of a nonlinear convective contribution indicates that, at this stage, the wave speed is not yet strongly amplitude dependent, and the profile evolution remains close to linear advection with weak dispersion. In this sense, the discovered form can be understood as a linearized KdV-type description, which is consistent with the standard picture of shallow-water long waves before strong steepening develops.

For the pre-breaking stage, the discovered equation becomes:

$$\eta_t^* + 0.90\eta_x^* + 4.75 \times 10^{-4}\eta_{xxx}^* + 4.31 \times 10^{-2}\eta^*\eta_x^* = 0, \tag{19}$$

The competitive optimization and model selection process is illustrated in Fig. 5d and Fig. S7b. The appearance of the nonlinear term ($\eta^*\eta_x^*$) is the key physical change. This term encodes amplitude-dependent advection, meaning that higher parts of the wave travel faster than lower parts, thereby sharpening the front and driving the system toward breaking. The dispersive term remains active, and its larger magnitude suggests that dispersive effects become more pronounced as the profile steepens. The recovered equation therefore captures a pre-breaking regime governed by the competition between nonlinear steepening and dispersive spreading, which is precisely the mechanism classically associated with KdV-type wave evolution in shallow water. From Fig. S7c, it can be found that the left-hand side term and right-hand side terms of the discovered PDEs for both pre-steepening and pre-breaking stage align well, which confirms the ability of these PDEs to describe the underlying physical process. These results demonstrate that MCO-PDE can recover stage-dependent wave equations directly from multiple-source experimental data, while preserving their physical interpretability and consistency with the observed wave evolution.

Compared with the equation previously discovered from the full evolution[20], which contains the high order compacton term $[(\eta^*)^2]_{xxx}$, the stage-related PDEs identified here have simpler terms. This difference arises from the scope of the regression. The previous formulation uses a single high order expression to represent the transition from weakly nonlinear propagation to strong wave steepening. In the small-amplitude limit, the compacton contribution becomes a higher order correction, so the equation reduces to the pre-steepening form obtained here. Under a dominant-wavenumber projection, the compacton term can be mapped onto an effective $\eta^*\eta_x^*$ contribution, yielding a KdV-type form consistent with the pre-breaking equation. The difference in fitted coefficients can be attributed to the correlation between the compacton term and the derivative bases $\eta^*$ and $\eta_{xxx}^*$, which allows part of the full-stage compensation to be absorbed by the apparent linear propagation and dispersive coefficients. Importantly, the previously discovered formulation assigns dataset-dependent coefficients, whereas our approach identifies a unified governing equation shared across datasets. Details of the analysis is provided in Supplementary Information S1.2.

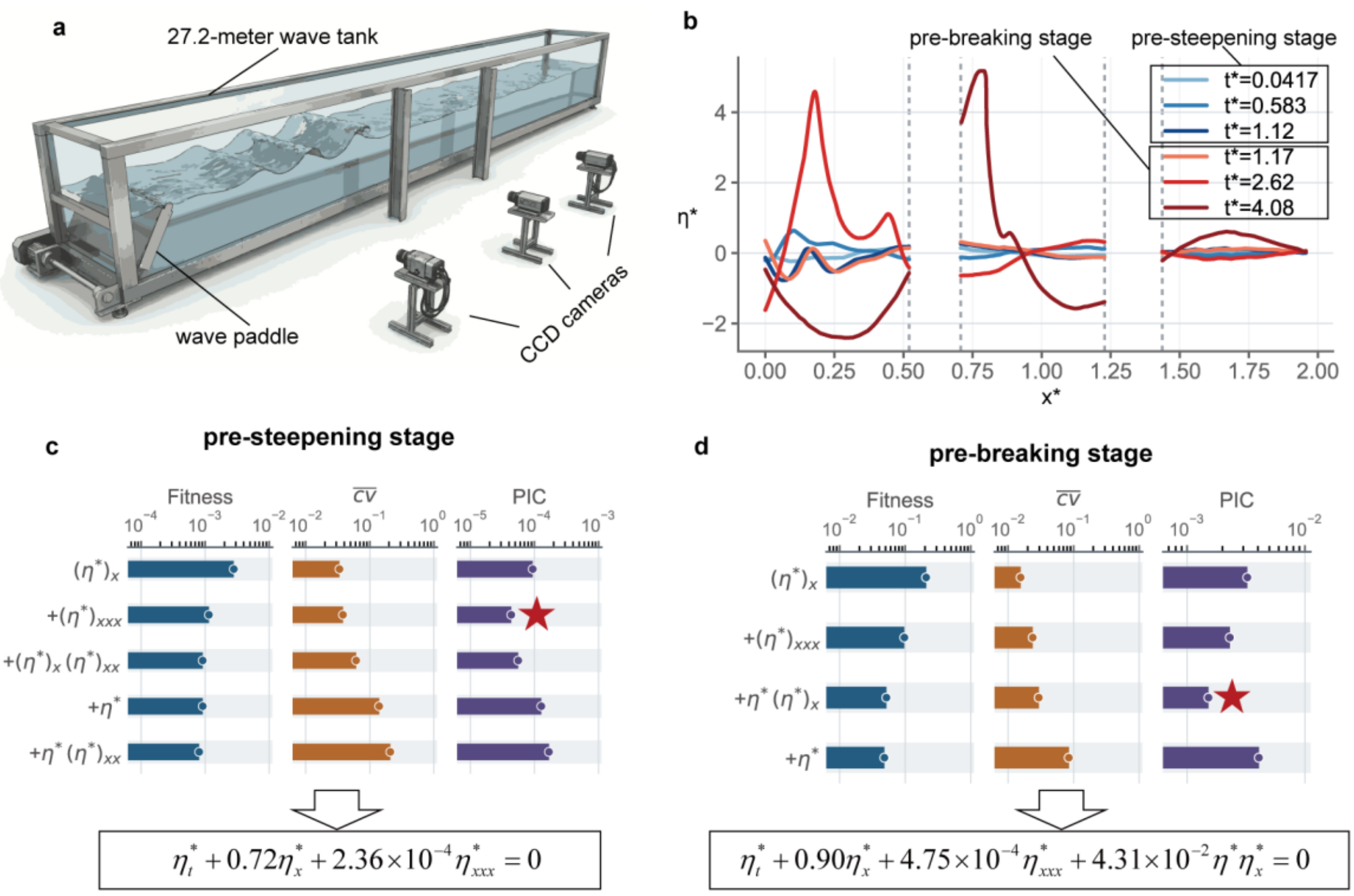

**Fig. 5. Discovery of stage-dependent wave equations from multi-source real-world experimental data. (a)**, Schematic of the 27.2-m wave tank equipped with a wave paddle and CCD camera arrays for measuring free-surface elevation. **(b)**, Normalized wave profiles at representative times, showing two distinct stages of evolution: a pre-steepening stage with mild nonlinear deformation and a pre-breaking stage with pronounced front steepening. **(c),** Model selection for the pre-steepening stage and the discovered PDE. **(d)**, Model selection for the pre-breaking stage and the discovered PDE.

## Discussion

In this work, we have presented a framework termed MCO-PDE for discovering governing PDEs by jointly learning from multiple datasets through competitive optimization. The central contribution of this work lies in the role of competitive optimization in enabling effective knowledge aggregation from multiple datasets. The ability to recover a governing equation from multi-source observations depends not simply on the number of datasets available, but on how much reliable information can be extracted and consolidated across them. In practice, different cases often vary substantially in both data quality and information content. Treating all cases as equally informative therefore weakens the fusion process. Our framework addresses this problem by introducing a soft-weighting strategy that assigns each case a dynamic confidence score and continuously updates these scores during optimization, thereby shifting the training emphasis towards the most informative cases. This departs from the implicit assumption made in most existing approaches that all datasets should contribute equally to discovery. As a result, MCO-PDE can make fuller use of the informative portion of a multi-source collection while suppressing the influence of less reliable cases. The supplementary experiments further show that, when inconsistent or corrupted cases are present, the soft-weighting mechanism automatically reduces their contributions during training, and such cases can be readily identified and excluded using a simple threshold on the learned weights (Supplementary Information S1.3).

Moreover, the proposed competitive optimization framework unifies knowledge discovery and knowledge embedding. For each candidate equation structure generated during the evolutionary search, the neural network estimates case-specific coefficients, aggregates them through competitive weighting into global coefficients, and then uses these global coefficients to constrain the surrogate models for all cases. This creates a self-reinforcing interaction between discovery and embedding. As the surrogate models improve, the estimated derivatives and regression coefficients become more reliable; in turn, the resulting global coefficients provide stronger and more coherent physical constraints, which further improve the surrogates. This improves the robustness of the MCO-PDE on limited data and high levels of noise. Another practical advantage of the framework is that it reduces the dependence on delicate case-by-case tuning that is common in conventional PDE discovery. Existing methods often rely on manually adjusted hyperparameters to balance equation simplicity and accuracy, and their performance can vary across datasets and noise levels. By contrast, MCO-PDE evaluates candidate structures through their global loss over all datasets and automatically redistributes weights across cases during optimization, which reduces the need for such manual balancing. Because a candidate equation must remain consistently supported across multiple datasets to achieve a low global loss, redundant terms and incorrect terms are less likely to persist.

The numerical sample efficiency of MCO-PDE under data-sparse constraints highlights the utility of multi-dataset fusion for scientific discovery. Even for high-dimensional systems, irregular domains, and heterogeneous PDEs, the aggregation of information across datasets enables correct equations to be identified with far fewer observations per case than are typically required by single-dataset discovery methods. This characteristic is advantageous for practical applications where continuous dense sensing is technically constrained or cost-prohibitive, yet repeated observations under varying configurations are attainable. The real-world wave-tank experiments provide an example. Despite segmented observations across multiple trials, MCO-PDE reconstructed physically interpretable, regime-dependent wave equations. More broadly, these results suggest that multi-source PDE discovery is especially well suited to experimental and engineering settings where observation is limited, replication is possible, and the underlying physics is shared across cases.

Several limitations should also be acknowledged. First, the current framework is designed for settings in which multiple datasets are governed by a shared equation structure and a common set of coefficients, up to estimation error. This assumption is appropriate for many repeated-observation scenarios, but it restricts the applicability of the method when different subsets of data arise from distinct physical regimes or are governed by different equations. Extending the framework to such settings will likely require an additional mechanism for identifying latent groups of datasets and assigning them to different governing structures before joint discovery can be carried out. Second, the present framework remains computationally demanding. For each candidate equation structure proposed by the genetic algorithm, the competitive optimization is utilized to evaluate its cross-dataset consistency and estimate the corresponding global coefficients. As a result, discovering a governing equation typically requires from tens of minutes to several hours, depending on the complexity of the gene library and the number of datasets. Notably, compared with strategies that perform separate surrogate training and equation discovery for every individual case, such as PINN-SR method, MCO-PDE already improves efficiency by carrying out unified optimization across datasets rather than repeating the entire procedure case by case. Despite these limitations, the proposed framework provides a promising path for extracting governing laws from heterogeneous multi-source data.

## Methods

### Neural surrogate training and derivative computation

Consider $S$ datasets $\{D^{(s)}\}_{s=1,\dots,S}$, each consisting of spatiotemporal observations of a field variable $u(\mathbf{x}, t)$ sampled under different initial conditions, boundary geometries, or forcing scenarios, but governed by the same unknown PDE. For each dataset $D^{(s)}$, we train an independent fully connected neural network $f_{\theta s}$ to approximate the field $u^{(s)}$ by minimizing a data-fidelity loss:

$$\mathscr{L}_{\text{data}}{}^{(s)} = \|f_{\theta s}(\mathbf{x}, t) - u^{(s)}\|^2 \tag{20}$$

In this work, we utilized a five-layer artificial neural network with 50 neurons in each hidden layer. The activation function is Sin, the optimizer is Adam with the learning rate of $10^{-3}$. The maximum training epoch is 5000, and the validation dataset is utilized for performing early stopping. Once trained, the neural network serves as a smooth surrogate from which arbitrary-order spatial and temporal derivatives can be computed via automatic differentiation. This avoids the ill-conditioning inherent in finite-difference approximations on sparse grids and provides noise-filtered derivative estimates at any collocation point, including points not present in the original dataset. The derivative fields $\{u, u_{\mathbf{x}}, u_{\mathbf{xx}}, u_{\mathbf{xxx}}, \dots\}$ constitute the basic gene set from which candidate equation terms are constructed.

### Competitive optimization and global coefficient estimation

For each candidate equation structure generated by the genetic search, we performed joint optimization across all datasets to determine whether a single set of coefficients could consistently explain the observed dynamics. The key idea is to estimate case-wise coefficients from each dataset, quantify their reliability, and then aggregate them into a global coefficient vector that serves as the shared physical constraint during training. This design allows datasets with stronger physical identifiability to contribute more strongly to equation discovery, while reducing the influence of less informative or less stable cases.

At each training epoch, all surrogates were optimized simultaneously with Adam using a learning rate of $10^{-3}$. For each case $s$, the observed samples contributed a data-fitting term, and an additional set of collocation points was drawn uniformly from the spatio-temporal domain of that case. Unless otherwise stated, $n_{colloc}$=1000 collocation points were used per case at each epoch. Automatic differentiation was then used to compute the left-hand-side temporal derivative and the spatial derivatives required by the candidate structure. For a candidate library with $K$ active terms, these quantities define a design matrix $\Phi_s \in \mathbb{R}^{n_{colloc} \times K}$, while the target vector is given by the corresponding temporal derivative on the left-hand side. In the present implementation, the left-hand side can be specified as either a first-order or second-order time derivative, depending on the target equation. To improve numerical stability, each column of $\Phi_s$ was scaled by its root-mean-square magnitude, without centring. The collocation set was then partitioned into a support subset and a query subset, with support fraction 0.7 by default. Coefficients were estimated from the support set by least squares, and then mapped back to the original scale. The least-squares regression on the support subset can be written as: $\hat{\beta}_s = \arg\min_{\beta} \|\Phi_s^{sup}\beta - y_s^{sup}\|_2^2$, where $\Phi_s^{sup}$ and $y_s^{sup}$ are the right-hand side terms and left-hand side term on the support subset. The query set was used only for reliability assessment and not for coefficient fitting. This separation is important because it suppresses structures that appear favourable only through incidental agreement on a particular collocation sample.

The reliability of each case was quantified through a composite score that combines predictive consistency and numerical identifiability. Specifically, we computed the normalized query error:

$$r_s = \frac{\|\mathbf{y}_s^{qry} - \hat{\boldsymbol{\Phi}}_s^{qry}\hat{\beta}_s\|^2}{\|\mathbf{y}_s^{qry}\|^2 + \varepsilon}, \tag{21}$$

together with the condition number of the support design matrix,

$$\kappa_s = \frac{\sigma_{\max}(\Phi_s^{\mathrm{sup}})}{\sigma_{\min}(\Phi_s^{\mathrm{sup}}) + \varepsilon}, \tag{22}$$

where $\sigma_{\max}$ and $\sigma_{\min}$ denote the largest and smallest singular values, respectively. These two quantities were combined as:

$$J_s = r_s + \lambda_\kappa \log(\kappa_s), \tag{23}$$

with $\lambda_\kappa = 10^{-2}$. Cases with lower scores are therefore preferred, as they correspond simultaneously to smaller held-out residuals and better-conditioned coefficient estimation.

The case scores were converted into soft competitive weights through

$$w_s = \frac{exp(-\tau J_s)}{\sum_j exp(-\tau J_j)}. \quad (24)$$

The parameter $\tau$ controls the sharpness of the competition. In our implementation, $\tau$ was increased linearly from 0.1 to 5.0 over the first one-third of the training epochs and then held fixed. Early in training, when derivative estimates are still unstable, this schedule yields a relatively broad consensus across datasets. As training proceeds, the weighting becomes increasingly selective, so that the global estimate is driven more strongly by the most reliable cases. The resulting global coefficient vector was defined as a weighted average,

$$\beta_{global} = \sum_s w_s \hat{\beta}_s. \quad (25)$$

This coefficient vector should therefore be understood as a cross-dataset consensus estimate rather than the fit from any single case. As the reliability-weighted consensus coefficient vector across datasets, and further introduce $\beta_{EMA}$ as its temporally smoothed version. Specifically, $\beta_{EMA}$ is initialized by the first available global estimate and then updated by exponential moving average,

$$\beta_{EMA}^{(1)} = \beta_{global}^{(1)},\ \beta_{\mathrm{EMA}}^{(e)} = (1-\rho)\beta_{\mathrm{EMA}}^{(e-1)} + \rho\beta_{\mathrm{global}}^{(e)}, \quad (26)$$

where $e$ denotes the training epoch and $\rho$=0.1 controls the smoothing rate. In this way, $\beta_{EMA}$ represents a stabilized cross-dataset coefficient estimate that suppresses transient oscillations while retaining the dominant trend of the optimization. The smoothed coefficients are then used to define the PDE residual for every case,

$$\mathcal{R}_s = y_s - \Phi_s \beta_{EMA}, \quad (27)$$

and the PDE loss was written as:

$$\mathcal{L}_{PDE} = \sum_s \| \mathcal{R}_s \|_2^2. \quad (28)$$

Importantly, $\beta_{EMA}$ entered this step as a detached constant, so that coefficient aggregation was not back-propagated through the surrogate networks. This stop-gradient treatment stabilizes optimization by preventing transient coefficient fluctuations from directly perturbing the field approximation.

The total training objective combined the data-fitting and PDE terms,

$$\mathcal{L} = \mathcal{L}_{data} + \alpha\, \mathcal{L}_{PDE}, \quad (29)$$

where $\mathcal{L}_{data}$ is the sum of mean-squared errors over the observed data from all cases. The PDE weight $\alpha$ was linearly increased from 0 to 1 over the first one-third of the training epochs and then held fixed. This warm-up strategy ensures that training is initially dominated by data fidelity, allowing the surrogates to form stable field representations before the shared PDE constraint is enforced strongly. As $\alpha$ and $\tau$ increase, the optimization progressively shifts from case-wise fitting towards cross-dataset physical agreement, and the case-wise coefficient estimates are driven towards a common solution. In this way, competitive optimization converts multi-dataset PDE discovery into a reliability-weighted consensus problem. Rather than treating all datasets as equally informative, the framework dynamically emphasizes those cases that provide the most stable and transferable physical signal, while still using all cases to refine the shared surrogate-constrained equation. The final output of this procedure is the smoothed global coefficient vector together with the associated data and PDE losses, which are subsequently used to evaluate the candidate structure within the evolutionary search.

**Genetic algorithm for equation-structure search**

To identify the governing equation structure shared across multiple datasets, we used a genetic algorithm to search over a symbolic library of candidate PDE terms. Each candidate equation was encoded as a genome, in which each module represented one candidate term assembled from the predefined gene library, for example field variables and their spatial derivatives. A genome therefore corresponded to a candidate right-hand-side structure for the target PDE, whereas the coefficients of the selected terms were not fixed by the genetic algorithm itself, but were estimated subsequently by the competitive optimization neural network during loss evaluation. This separation between structure search and coefficient estimation allowed the evolutionary process to focus on discovering the most plausible term combinations while delegating parameter fitting to the multi-case surrogate model.

The search started from a randomly initialized population of genomes generated from the prescribed basic gene library. Each candidate genome was converted into a term library and evaluated by training the competitive optimization neural network across all datasets. For a given structure, the network simultaneously fitted the surrogate models, estimated the case-wise coefficients, aggregated them into global coefficients, and returned both the data loss and the physics loss. We defined the loss of a candidate structure as the product of these two quantities,

$$\text{Loss} = \mathcal{L}_{data} \times \mathcal{L}_{pde}, \tag{30}$$

so that a structure was favored only when it explained the observations accurately and remained consistent with the governing physics across datasets. This multiplicative form was used to penalize structures that performed well in only one aspect, for example fitting the data while failing to satisfy the PDE constraint, or vice versa. In all loss evaluations, collocation-based PDE training used 1,000 collocation points by default.

To improve efficiency, we cached the loss of previously evaluated genomes using a canonical genome key obtained by sorting modules and module contents. This avoided repeated training when the same structure reappeared in later generations under a different syntactic ordering. Within each generation, all candidate genomes were ranked according to their loss values, and the best-performing structures were retained as elites. The elite fraction was set to 10% of the population, with at least one elite preserved in every generation. These elite genomes were copied directly into the next generation to stabilize inheritance of the best structures found so far. The remaining members of the new population were generated by combining exploitation and exploration. First, additional genomes were sampled de novo from the gene library. Then, each sampled genome underwent crossover with a randomly chosen elite genome, followed by mutation and deduplication. Crossover was implemented by joining two parental genomes at randomly chosen cut points, thereby exchanging partial term sets between structures. Mutation then randomly performed one of three operations: deleting an existing module, adding a newly sampled module, or replacing one module with a newly sampled one. This strategy allowed the search to preserve well-performing substructures while continuously injecting new structural variations. After these operations, genomes were again sanitized before entering the next generation. The evolutionary process continues until the maximum number of generations is reached, after which the best-performing structure is selected. In this work, the population size is 50 and the maximum generation is 10 by default.

**Model selection via physics-informed information criterion**

After the genetic search, model selection was performed in a second stage to identify the most parsimonious and physically consistent substructure within the best candidate genome. Specifically, we first extracted the top-ranked genome from the cached genetic-search results, retrained the multi-case competitive model for this full structure, and then re-estimated the coefficients of all active terms separately for each case using 1,000 newly sampled collocation points per dataset. For each term $k$, we collected the coefficient estimates across cases, computed their mean $\mu_k$ and standard deviation $\sigma_k$, and quantified their cross-dataset stability by the coefficient of variation

$$\mathrm{CV}_k = \frac{\sigma_k}{|\mu_k| + \varepsilon}, \tag{31}$$

where $\varepsilon$ is a small numerical constant. Terms with smaller $\mathrm{CV}_k$ were regarded as more stable across datasets and therefore more likely to reflect shared physics rather than case-specific fluctuations. The terms in the selected genome were then sorted in ascending order of $\mathrm{CV}_k$, and a sequence of nested candidate submodels was constructed by adding terms one by one according to this ranking. Thus, the first candidate contained only the most stable term, the second candidate contained the two most stable terms, and so forth until the full genome was recovered. For each submodel, we re-ran the competitive optimization procedure for 1,000 epochs and recorded the final data loss and PDE loss. As in the genetic-search stage, these two quantities were combined into a single loss measure, as defined in Eq. (30), so that a candidate submodel was favored only when it achieved both accurate data reconstruction and good cross-dataset physical consistency.

To balance goodness of fit against structural compactness and coefficient stability, we defined a physics-informed information criterion based on the mean coefficient of variation of the retained terms,

$$\overline{\mathrm{CV}} = \frac{1}{m}\sum_{k=1}^{m} \mathrm{CV}_k, \tag{32}$$

where $m$ is the number of terms in the current submodel, and then combined it with the loss as:

$$\mathrm{PIC} = \mathrm{Loss} \times \overline{\mathrm{CV}}. \tag{33}$$

This criterion favors submodels that simultaneously satisfy three requirements: low physical residual, good agreement with the observed data, and stable coefficient estimates across datasets. For a single dataset, instead of comparing coefficients across cases, we partitioned the spatial domain into five local windows, generated meta-data within each window over the full temporal range, and re-estimated the coefficients accordingly. The mean coefficient of variation across these local windows was then used to rank the terms and to compute the PIC[29].

**Acknowledgements**

This work was supported and partially funded by the National Natural Science Foundation of China (Grant 52288101, 12501744, 12572266), the China National Postdoctoral Program for Innovative Talents (Grant No. BX20250063), the China Postdoctoral Science Foundation (Grant No. 2024M761535), the National Key Research and Development Program (2024YFF1500600), and the Yongjiang Talent Program of Ningbo (2022A-242-G). This work is supported by the High Performance Computing Centers at Eastern Institute of Technology, Ningbo, and Ningbo Institute of Digital Twin.

**Author contributions**

H. X., S. L., Y. C. and D. Z. conceived the idea, designed the study, and analyzed the results. H. X. developed the algorithm, performed the computations, and generated the results and figures. H. X., Y. C. and D. Z. wrote and edited the manuscript. D. Z. supervised the entire project.

**Declaration of interests**

The authors declare that they have no competing interests.

**Data availability**

The dataset generated in this study has been deposited in the GitHub repository
https://github.com/woshixuhao/multi_source_PDE_discovery/tree/main/data

**Code availability**

All of the original code has been deposited at the website
https://github.com/woshixuhao/multi_source_PDE_discovery/tree/main/code

**References**

1. Brunton, S. L. & Kutz, J. N. Promising directions of machine learning for partial differential equations. *Nat. Comput. Sci.* **4**, 483–494 (2024).
2. Brunton, S. L., Proctor, J. L., Kutz, J. N. & Bialek, W. Discovering governing equations from data by sparse identification of nonlinear dynamical systems. *Proc. Natl. Acad. Sci. U. S. A.* **113**, 3932–3937 (2016).
3. Rudy, S. H., Brunton, S. L., Proctor, J. L. & Kutz, J. N. Data-driven discovery of partial differential equations. *Sci. Adv.* **3**, 1–7 (2017).
4. Messenger, D. A. & Bortz, D. M. Weak SINDy for partial differential equations. *J. Comput. Phys.* **443**, 110525 (2021).
5. Tang, M., Liao, W., Kuske, R. & Kang, S. H. WeakIdent: Weak formulation for identifying differential equation using narrow-fit and trimming. *J. Comput. Phys.* **483**, 112069 (2023).
6. Stephany, R. & Earls, C. Weak-PDE-LEARN: A weak form based approach to discovering PDEs from noisy, limited data. *J. Comput. Phys.* **506**, 112950 (2024).
7. Xu, H., Zhang, D. & Wang, N. Deep-learning based discovery of partial differential equations in integral form from sparse and noisy data. *J. Comput. Phys.* **445**, (2021).
8. Xu, H., Chang, H. & Zhang, D. DLGA-PDE: Discovery of PDEs with incomplete candidate library via combination of deep learning and genetic algorithm. *J. Comput. Phys.* **418**, 109584 (2020).
9. Maslyaev, M., Hvatov, A. & Kalyuzhnaya, A. Data-Driven Partial Derivative Equations Discovery with Evolutionary Approach. in *Computational Science -- ICCS 2019* (eds. Rodrigues, J. M. F. et al.) 635–641 (Springer International Publishing, Cham, 2019).

10. Du, M., Chen, Y. & Zhang, D. DISCOVER: Deep identification of symbolically concise open-form partial differential equations via enhanced reinforcement learning. *Phys. Rev. Res.* **6**, (2024).
11. Chen, Y., Luo, Y., Liu, Q., Xu, H. & Zhang, D. Symbolic genetic algorithm for discovering open-form partial differential equations (SGA-PDE). *Phys. Rev. Res.* **4**, (2022).
12. Petersen, B. K. *et al.* Deep Symbolic Regression: Recovering Mathematical Expressions From Data Via Risk-Seeking Policy Gradients. *ICLR 2021 - 9th International Conference on Learning Representations* (2021).
13. Long, Z., Lu, Y., Ma, X. & Dong, B. Pde-net: Learning pdes from data. in *International conference on machine learning* 3208–3216 (PMLR, 2018).
14. Long, Z., Lu, Y. & Dong, B. PDE-Net 2.0: Learning PDEs from data with a numeric-symbolic hybrid deep network. *J. Comput. Phys.* **399**, 108925 (2019).
15. Both, G. J., Choudhury, S., Sens, P. & Kusters, R. DeepMoD: Deep learning for model discovery in noisy data. *J. Comput. Phys.* **428**, 109985 (2021).
16. Xu, H., Chang, H. & Zhang, D. Dl-pde: Deep-learning based data-driven discovery of partial differential equations from discrete and noisy data. *Commun. Comput. Phys.* **29**, 698–728 (2021).
17. Xu, H. & Zhang, D. Robust discovery of partial differential equations in complex situations. *Phys. Rev. Res.* **3**, (2021).
18. Chen, Z., Liu, Y. & Sun, H. Physics-informed learning of governing equations from scarce data. *Nat. Commun.* **12**, 1–13 (2021).
19. Thanasutives, P., Morita, T., Numao, M. & Fukui, K. Noise-aware physics-informed machine learning for robust PDE discovery. *Mach. Learn. Sci. Technol.* **4**, 015009 (2023).
20. Xu, H. *et al.* Generative discovery of partial differential equations by learning from math handbooks. *Nat. Commun.* **16**, 10255 (2025).
21. Du, M., Chen, Y., Wang, Z., Nie, L. & Zhang, D. LLM4ED: Large Language Models for Automatic Equation Discovery. *arXiv preprint arXiv:2405.07761* (2024).
22. Shojaee, P., Meidani, K., Gupta, S., Farimani, A. B. & Reddy, C. K. Llm-sr: Scientific equation discovery via programming with large language models. *arXiv preprint arXiv:2404.18400* (2024).
23. Zanna, L. & Bolton, T. Data-Driven Equation Discovery of Ocean Mesoscale Closures. *Geophys. Res. Lett.* **47**, (2020).
24. Beetham, S., Fox, R. O. & Capecelatro, J. Sparse identification of multiphase turbulence closures for coupled fluid-particle flows. *J. Fluid Mech.* **914**, (2021).
25. Tod, G., Both, G.-J. & Kusters, R. Discovering PDEs from Multiple Experiments. http://arxiv.org/abs/2109.11939 (2021).
26. Castanedo, F. A review of data fusion techniques. *The scientific world journal* **2013**, 704504 (2013).
27. Peng, X. *et al.* Moment matching for multi-source domain adaptation. in *Proceedings of the IEEE/CVF international conference on computer vision* 1406–1415 (2019).
28. Meng, T., Jing, X., Yan, Z. & Pedrycz, W. A survey on machine learning for data fusion. *Information Fusion* **57**, 115–129 (2020).

29. Xu, H., Zeng, J. & Zhang, D. Discovery of Partial Differential Equations from Highly Noisy and Sparse Data with Physics-Informed Information Criterion. *Research* **6**, (2023).
30. Stephany, R. & Earls, C. PDE-READ: Human-readable partial differential equation discovery using deep learning. *Neural Networks* **154**, 360–382 (2022).
31. Gao, E.-H., Ge, C., Jiang, Y. & Zhou, Z.-H. Discovering Symbolic Partial Differential Equation by Abductive Learning. in *The Thirty-ninth Annual Conference on Neural Information Processing Systems*.
32. Meng, Y. & Qiu, Y. Sparse discovery of differential equations based on multi-fidelity Gaussian process. *J. Comput. Phys.* **523**, 113651 (2025).
33. Cao, R., Padilla, E. M. & Callaghan, A. H. The influence of bandwidth on the energetics of intermediate to deep water laboratory breaking waves. *J. Fluid Mech.* **971**, A11 (2023).
34. Cao, R., Padilla, E. M., Fang, Y. & Callaghan, A. H. Identification of the free surface for unidirectional nonbreaking water waves from side-view digital images. *IEEE Journal of Oceanic Engineering* (2024).

# **Supplementary Information** for **Joint discovery of governing partial differential equations from multi-source datasets by competitive optimization**

Hao Xu[1,2], Siyu Lou[1,3], Yuntian Chen[1,4,*], and Dongxiao Zhang[1,*]

[1] Zhejiang Key Laboratory of Industrial Intelligence and Digital Twin, Eastern Institute of Technology, Ningbo, Zhejiang 315200, China

[2] Department of Electrical Engineering, Tsinghua University, Beijing 100084, P. R. China.

[3] Shanghai Jiao Tong University, School of Computer Science, Shanghai, China

[4] Ningbo Institute of Digital Twin, Eastern Institute of Technology, Ningbo, Zhejiang 315200, P. R. China

[*] Corresponding authors.

Email address: ychen@eitech.edu.cn (Y. Chen); dzhang@eitech.edu.cn (D. Zhang).

1. **Supplementary text**

## 1.1 Datasets description

All synthetic datasets used in this work were generated from prescribed governing equations in MATLAB or Python under controlled initial and boundary conditions. Across all examples, the equation coefficients were fixed, whereas the multiple cases were created by varying the initial conditions, the computational domain, or both. This design ensured that all cases within a dataset family shared the same underlying physics while still exhibiting sufficiently diverse spatiotemporal patterns for multi-source discovery.

**Burgers' equation**

$$u_t = -uu_x + 0.1u_{xx} \quad \text{(S.1)}$$

The Burgers' dataset was generated on the periodic spatial domain $x\in[-8, 8]$ and time interval $t\in[0,10]$. The equation was solved with Chebfun's spin solver using a trigonometric representation, with $n_x$=256 spatial points, $n_t$=201 saved time snapshots, and internal time step $\Delta t = 2\times10^{-4}$. The exported solution array therefore has size 256 × 201 for each case. Seven cases were generated from six periodic initial conditions: case 1, $u_0(x) = \sin(3\pi x/8)$; case 2, $u_0(x) = \sin(\pi x/8) + 0.5 \sin(2\pi x/8)$; case 3, $u_0(x) = 0.8 \cos(\pi x/4)$; case 4, $u_0(x)= 0.6 \cos(\pi x/8) \sin(\pi x/2)$; case 5, $u_0(x) = 0.7 \sin(\pi x/8) + 0.25 \cos(3\pi x/8)$; case 6, $u_0(x)=0.9 [\sin(\pi x/8) +1/3 \sin(3\pi x/8) + 1/5 \sin(5\pi x/8)]$; case 7, $u_0(x) = \sin(\pi x/8)$.

**KdV equation**

$$u_t = -uu_x - 0.0025u_{xxx}, \quad \text{(S.2)}$$

The KdV dataset was generated on $x\in[-1, 1]$ and $t\in[0, 1]$. The solver used a custom compact finite-difference discretization for the first- and third-order spatial derivatives together with explicit Runge–Kutta time integration. The raw spatial grid contained 513 points including the duplicated periodic endpoint, and the exported solution removed this duplicate point, yielding $n_x$=512 spatial samples. Time integration used $\Delta t$=2.5×10$^{-6}$, giving 400001 raw time steps, and the solution was saved every 2000 steps, resulting in $n_t$=201 stored time snapshots. Seven initial conditions were used: case 1, a cosine wave $u_0(x)= \cos(\pi x)$; case 2, a single soliton with $A$=1.0 and $k$=8.0; case 3, a two-soliton superposition with $(A_1, k_1, x_1) = (1.0, 9.0, -0.35)$ and $(A_2, k_2, x_2) = (0.6, 13.0, 0.35)$; case 4, a cnoidal wave $u_0(x)= 0.2 + 0.8\, cn^2(kx; m)$ with $m$=0.9 and $k=2K(m)/L$; case 5, a multi-mode Fourier wave $0.35 \cos(\pi x) + 0.18 \cos(2\pi x) + 0.10 \sin(3\pi x)$; case 6, a Gaussian pulse with amplitude 0.9 and width $\sigma$=0.15; and case 7, a smoothed top-hat pulse with $A$=0.9, $w$=0.6 and $\delta$=0.03.

**Klein–Gordon equation**

$$u_{tt} = -5u+0.5u_{xx}, \quad \text{(S.3)}$$

The Klein–Gordon dataset was generated on $x\in[-1, 1]$ and $t\in[0, 3]$, with homogeneous Dirichlet boundary conditions at $x$=–1 and $x$=1. The equation was solved with Chebfun's chebop2 formulation, in which the spatial boundary conditions and the initial conditions $u(x, 0)=u_0(x)$ and $u_t(x, 0) =v_0(x)$ were imposed directly in the operator definition. The exported grid used $n_x$=201 and $n_t$ = 201. Seven cases were considered. Case 1 used $u_0(x)=\exp(-20x^2)$ and $v_0(x)$=0. Case 2 used a left-localized Gaussian wavepacket with zero initial velocity. Case 3 used a right-localized multi-frequency Gaussian packet with zero initial velocity. Case 4 used a Gaussian-modulated wavepacket centered at $x_0$=0.15, again with zero initial velocity. Case 5 used two antisymmetric Gaussian pulses with $A$=0.7, $w$=60, $x_1$=–0.35 and $x_2$=0.35, and $v_0$= 0. Case 6 used a multi-mode standing-wave mixture $u_0(x)=0.8\sin(\pi(x + 1)/2) + 0.3 \sin(3\pi(x + 1)/2)$, with $v_0$=0. Case 7 used a Gaussian packet with matched nonzero initial velocity, where both $u_0$ and $v_0$ were localized sinusoidal wavepackets of opposite sign.

**Allen–Cahn equation**

$$u_t = u - u^3 + 0.003u_{xx}, \tag{S.4}$$

The Allen–Cahn dataset was generated on $x\in[-1, 1]$ using Chebfun's spin solver with $n_x$=256 spatial points and internal time step $\Delta t$=0.001. The saved output interval was fixed at 0.05, but the final time varied by case in order to produce nontrivial dynamics over different temporal ranges. Cases 2 and 3 were saved on $t\in[0, 3]$, cases 4 and 5 on $t\in[0, 5]$, and cases 1, 6 and 7 on $t\in[0, 10]$. This yields $n_t$=61, 101 and 201 saved snapshots, respectively. The initial conditions were: case 1, $u_0(x)= 0.2\sin^5(2\pi x)+0.8\sin(5\pi x)$; case 2, $u_0(x)=0.50 \cos(\pi x)+0.28 \sin(2\pi x)+0.18\cos(3\pi x)$; case 3, a Gaussian-modulated sine packet $0.80 \exp(-(x+0.15)^2/0.18)\sin(4\pi x)$; case 4, $u_0(x)=0.55\sin(\pi x)+0.20\sin(2\pi x)$; case 5, $u_0(x)=0.75 \exp(-x^2/(2 \times 0.35^2))\cos(3\pi x)$; case 6, $u_0(x)=0.25\cos(3\pi x)$; and case 7, a reproducible random low-frequency Fourier mixture generated with fixed random seed 1.

**Two-dimensional heat equation on irregular domains**

$$u_t=0.2u_{xx}+0.2u_{yy}. \tag{S.5}$$

The two-dimensional diffusion dataset was generated on geometry-adapted unstructured finite-element meshes using MATLAB PDE Toolbox with quadratic elements. The exported data were packed as tuples ($x$, $y$, $t$, $u$), so no fixed $n_x$ and $n_y$ exist across cases; only the number of saved time snapshots was fixed, with $n_t$=50 for every case. If a solution became too close to steady state, the terminal time was reduced and the problem was re-solved while keeping the same number of snapshots. The seven cases differed in both geometry and boundary conditions. Case 1 used a rectangle $[0,1]\times[0, 0.8]$ with homogeneous Dirichlet boundary conditions. Case 2 used a triangle with vertices (0, 0), (1, 0) and (0.2, 0.9) and homogeneous Neumann boundary conditions. Case 3 used a unit disk with Dirichlet boundaries. Case 4 used an annulus with outer radius 1.0 and inner radius 0.35 and Dirichlet boundaries. Case 5 used an L-shaped domain with Neumann boundaries. Case 6 used a gear-shaped radial domain with Dirichlet boundaries. Case 7 used a smoothed heart-shaped domain with Neumann boundaries. The corresponding initial conditions were defined as mixed sine modes on case-specific reference boxes.

**Two-dimensional contaminant transport equation**

$$u_t = -0.2u_x - 0.15u_y + 0.05u_{xx} + 0.05u_{yy}, \tag{S.6}$$

The contaminant transport dataset was generated on the rectangular domain $x\in[0, 1.5]$, $y\in[0, 1]$ with homogeneous Dirichlet boundary conditions on all boundaries. A regular Cartesian grid with $n_x$=101 and $n_y$=101 was used. The time step was chosen from the minimum of the diffusion stability limit and the advection CFL limit, giving 2889 raw time steps over $t\in[0, 1]$. The solution was advanced explicitly, using central differences for diffusion and a second-order TVD MUSCL scheme with the minmod limiter for the advection term. Every 20th time step was saved, together with the final step, yielding $n_t$=146 stored snapshots. Each initial condition was scaled by an additional factor of 2.0 before time integration. The seven cases corresponded to seven source geometries. Case 1 was a single isotropic Gaussian centered at (0.30, 0.35). Case 2 was an anisotropic Gaussian centered at (0.45, 0.28). Case 3 was a smooth rectangular patch constructed from tanh transitions, centered at (0.80, 0.45) Case 4 contained two Gaussian sources located at (0.40, 0.55) and (0.60, 0.35). Case 5 was a ring-shaped source centered at (0.85, 0.40). Cases 6 and 7 were generated from multi-source interior Gaussian configurations.

**Three-dimensional heat equation**

$$u_t=0.01u_{xx}+ 0.01u_{yy} + 0.01u_{zz}, \tag{S.7}$$

The three-dimensional diffusion dataset was generated in the cuboid [0,1]×[0,1]×[0,1] with homogeneous Neumann boundary conditions on all six faces. The computational grid used $n_x=n_y=n_z=41$. Time integration was carried out by an explicit finite-difference scheme with CFL=0.45, giving $\Delta t$ = 0.009375 and 214 raw time steps over $t\in[0, 2]$. Every tenth step was saved, together with the final step, so that the exported dataset contains $n_t$=23 saved time snapshots. The seven cases were defined by different smooth mixed-mode initial conditions. Specifically, cases 1–7 used: mode(2, 2, 2) + 0.35 mode(1, 3, 2); mode(2, 1, 3) + 0.35 mode(1, 3, 2); mode(2, 2, 3) + 0.35 mode(1, 3, 2); mode(2, 2, 2) + 0.30 mode(1, 3, 3); mode(3, 2, 2) + 0.35 mode(1, 3, 2); mode(2, 3, 2) + 0.35 mode(1, 2, 3); and mode(2, 2, 2) + 0.25 mode(3, 1, 2). After construction, a constant offset of 2.0 was added to the initial field, which shifts the mean level without changing the second-derivative structure of the solution.

## 1.2 Analysis of the stage-dependent wave equations and their relation to the previously discovered full-stage equation

The stage-dependent wave equations identified in this work differ from the full-stage equation previously obtained by EqGPT[20]in both structural form and coefficient assignment. The EqGPT formulation represents the whole wave evolution with a high-order compacton term, $Q=[(\eta^*)^2]_{xxx}$, and assigns dataset-dependent coefficients to different wave cases. In contrast, the present stage-dependent analysis identifies simpler governing forms that are shared across datasets within each physical stage. The pre-steepening stage is described by

$$\eta_t+0.72\eta^*_x+2.36\times10^{-4}\eta^*_{xxx}=0, \tag{S.8}$$

while the pre-breaking stage is described by:

$$\eta^*_t+0.90\eta^*_x+4.75\times10^{-4}\eta^*_{xxx}+4.31\times10^{-2}\eta^*\eta^*_x=0, \tag{S.9}$$

These equations can be understood as stage-wise reductions of the full-stage EqGPT expression. For clarity, the full-stage equation discovered by EqGPT can be written in a signed form as:

$$\eta^*_t+c^{(m)}_x\eta^*_x+c^{(m)}_{xxx}\eta^*_{xxx}+c^{(m)}_Q Q=0, \tag{S.10}$$

where $m$ denotes the dataset and $c^{(m)}_x$, $c^{(m)}_{xxx}$, and $c^{(m)}_Q$ are case-specific coefficients. The structural difference between Eqs. S8–S9 and Eq. S10 is mainly associated with the compacton term $Q$. The compacton term can be expanded as

$$Q=\left[(\eta^*)^2\right]_{xxx}=2\eta^*\eta^*_{xxx}+6\eta^*_x\eta^*_{xx}. \tag{S.11}$$

This expansion shows that $Q$ contains both an amplitude-modulated dispersive component and a slope-curvature component. Hence, it can contribute to apparent propagation, dispersion, and nonlinear steepening when projected onto a lower-order basis.

In the small-amplitude regime, let $\eta^*=\epsilon h$, $0<\epsilon\ll 1$. Then $\eta^*_x=O(\epsilon)$, $\eta^*_{xxx}=O(\epsilon)$, while $Q=\left[(\eta^*)^2\right]_{xxx}=O(\epsilon^2)$. Thus, the compacton term enters as a higher-order correction in the pre-steepening stage. Keeping the leading-order terms in Eq. S3 gives:

$$\eta^*_t+c_x^{(m)}\eta^*_x+c_{xxx}^{(m)}\eta^*_{xxx}\sim 0, \tag{S.12}$$

which has the same structure as Eq. S8. This provides the asymptotic connection between the full-stage EqGPT equation and the stage-resolved pre-steepening equation. To validate this analysis, we further refitted the pre-steepening equation by adding the compacton term, and obtained

$$\eta_t+0.745\eta^*_x+2.78\times10^{-4}\eta^*_{xxx}-3.63\times10^{-5}Q=0. \tag{S.13}$$

Compared with Eq. S8, adding the compacton term changes the coefficient of the coefficient of $\eta^*_x$ changes by about 2.2% while reducing RMSE by 5.74%. These results indicates that the compacton term provides a minor residual correction in the pre-steepening regime, whereas the leading-order dynamics are already captured by the propagation and dispersive terms in Eq. S8. This is consistent with the asymptotic estimate in Eq. S12 that $Q$ can be omitted from the leading-order pre-steepening equation.

A second reduction can be derived for the pre-breaking stage. In a local narrow-band approximation with an effective wavenumber $k_{eff}$, the second derivative of a slowly modulated wave quantity $g$ satisfies $\partial_x^2 g\approx -k_{eff}^2 g$. Since $Q=\partial_x^2\left[(\eta^*)^2\right]_x$, one obtains $Q\approx\lambda\eta^*\eta^*_x$. Therefore, the compacton term in the full-stage EqGPT equation can reduce to a KdV-type nonlinear steepening term during the pre-breaking stage. This gives the structural basis for Eq. S9. To verify this interpretation, we refitted the EqGPT form in Eq. S10 using the pre-breaking data. The refitted equation is:

$$\eta_t+0.92\eta^*_x+5.77\times10^{-4}\eta^*_{xxx}-4.05\times10^{-5}Q=0. \tag{S.14}$$

The resulting RMSE is 0.778, which is nearly identical to that of the discovered KdV-type pre-breaking equation (RMSE=0.779; Eq. S9), representing a negligible relative difference of 0.13%. This close agreement indicates that the compacton term $Q$ and $\eta^*\eta^*_x$ provide nearly equivalent empirical descriptions of the pre-breaking dynamics. By contrast, when the pre-steepening governing structure, consisting only of $\eta^*_x$ and $\eta^*_{xxx}$, is refitted to the same pre-breaking data, the RMSE increases markedly to 0.942. This increased error suggests that propagation and dispersion alone are insufficient to describe the pre-breaking stage, where a nonlinear contribution is required to represent wave steepening.

### 1.3 Robustness to mismatched cases in multi-source discovery

A key assumption in multi-source PDE discovery is that the input datasets share a common underlying governing law. In practice, however, a dataset collection may contain mismatched cases arising from experimental failure or samples generated under a different physical mechanism. It is therefore important to test whether the proposed MCO-PDE framework can remain reliable when such corrupted cases are mixed into an otherwise consistent multi-source set. To examine this issue, we constructed a controlled robustness experiment in which six datasets were generated from the KdV equation, whereas one additional dataset was deliberately replaced by a mismatched case governed by a different pattern evolution (Fig. S8a). All seven cases were then treated as a single multi-source dataset and processed by MCO-PDE. Each case contained 1,000 sampled data points. The purpose of this experiment was to determine whether the competitive weighting mechanism could automatically suppress the inconsistent case and preserve the correct governing equation shared by the remaining majority.

The integrated fault-tolerance mechanism automatically excludes any data source from coefficient aggregation once its competitive weight drops below a prescribed threshold. Under this configuration, MCO-PDE accurately reconstructs the expected PDE, with the training trajectories exhibiting a distinct bifurcation between the consistent cases and the anomalous outlier (Fig. S8b-e). The six KdV datasets maintained relatively stable and comparable weights during optimization, whereas the mismatched case exhibited a rapid and persistent weight decay, eventually falling below the pruning threshold and being removed from the global aggregation step. After this exclusion, the estimated coefficients across the remaining six cases quickly became more consistent and converged towards a common solution. Several features of this experiment are noteworthy. First, the mismatched case was not removed a priori, but identified automatically during optimization through its low cross-dataset compatibility. Second, the weight evolution is highly interpretable. MCO-PDE first allows all cases to participate, then progressively suppresses the inconsistent one as the physical inconsistency becomes more evident. Third, once the mismatched case is excluded, both the coefficient trajectories and the overall loss curves become more stable. This property is particularly important for real applications, where multi-source datasets are rarely perfectly curated and may contain outliers or failed measurements. In such settings, MCO-PDE provides a practical safeguard that enables reliable equation discovery without requiring manual screening of every individual case.

## 2. Supplementary figures and tables

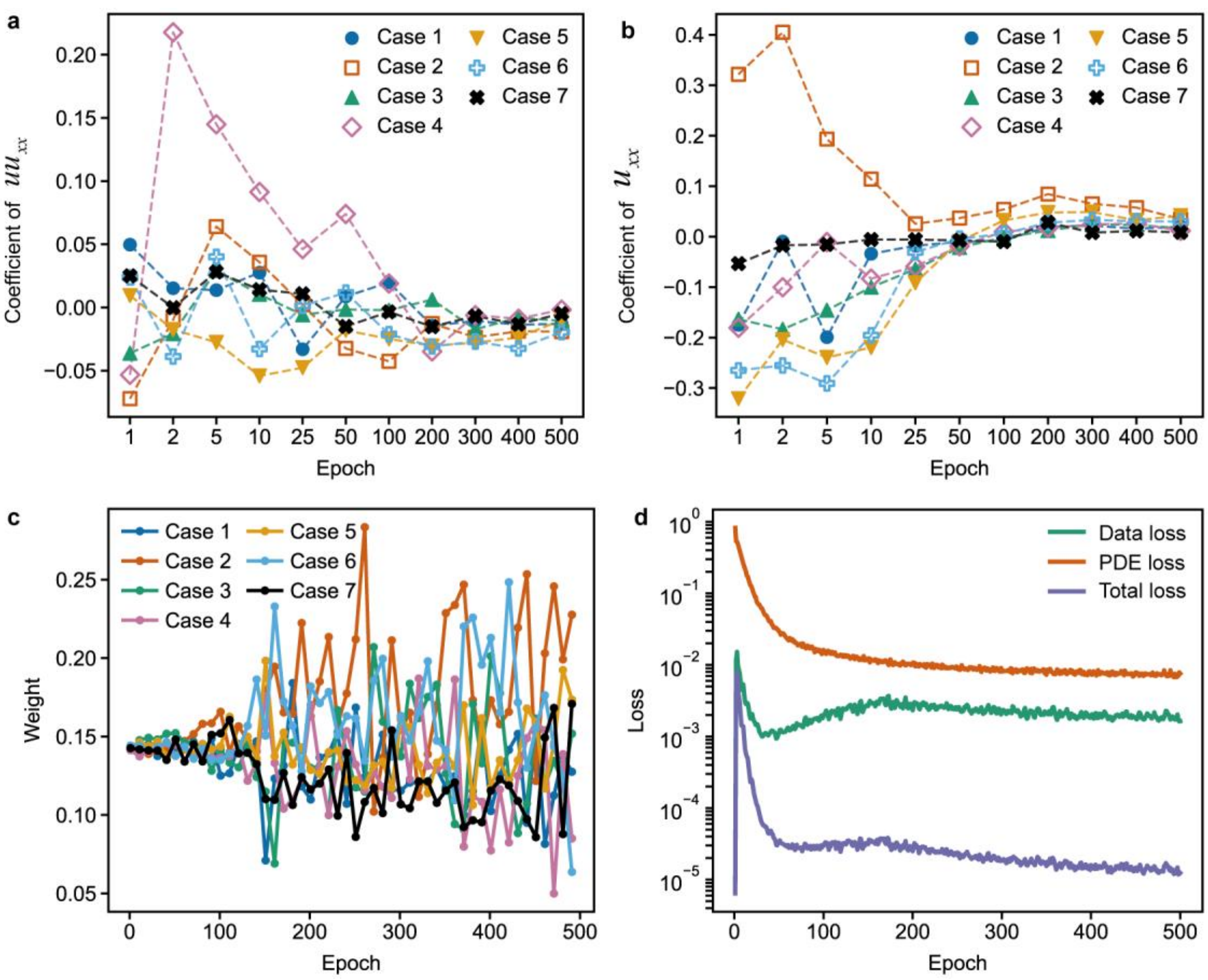

**Fig. S1. Training dynamics of multi-source discovery for the incorrect equation structure. (a,b),** Evolution of the identified coefficients for the term $uu_{xx}$ and $u_{xx}$ across the seven cases during training. **(c),** Evolution of the competitive weights assigned to each case. **(d),** Evolution of the data loss, PDE loss and total loss during optimization.

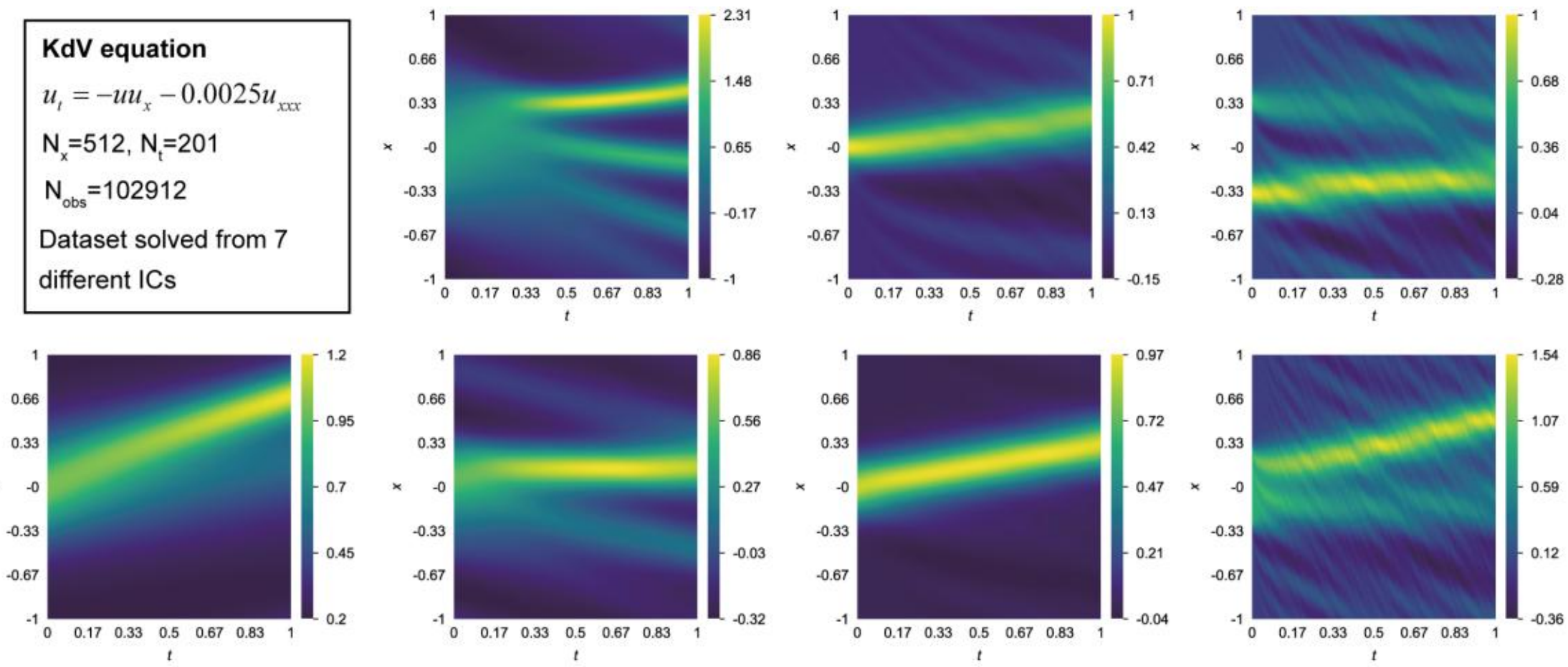

**Fig. S2. The multiple source dataset for discovering KdV equation.** These seven datasets are solved from different initial conditions (ICs).

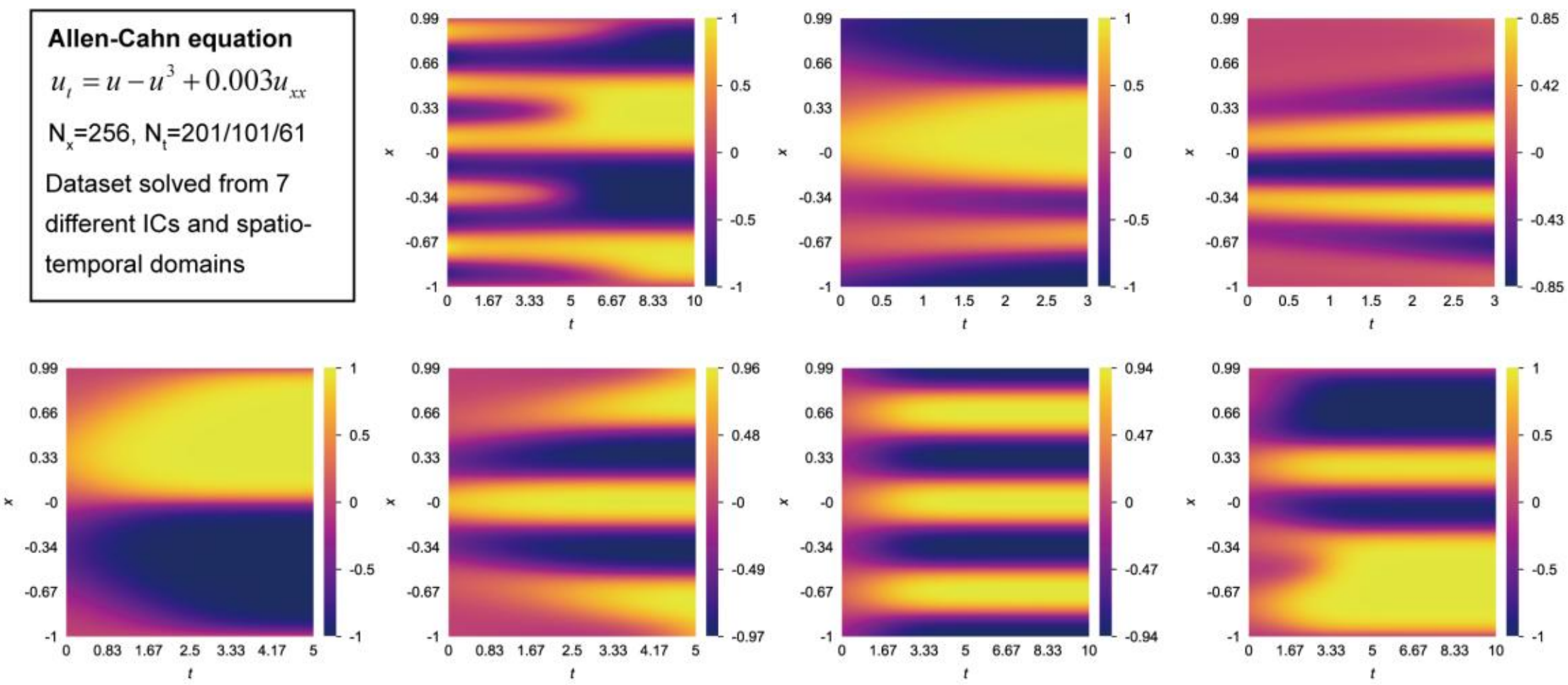

**Fig. S3. The multiple source dataset for discovering Allen-Cahn equation.** These seven datasets are solved from different initial conditions (ICs) and different spatio-temporal domains.

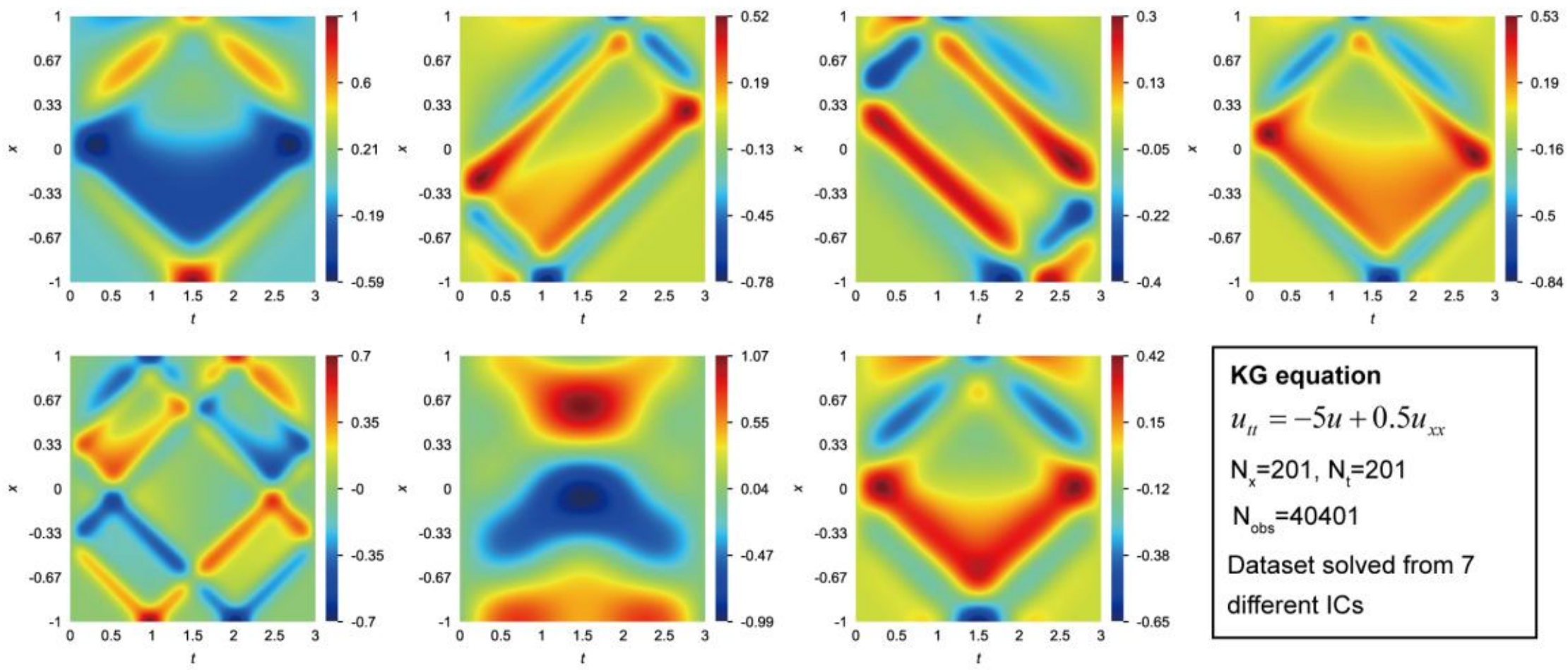

**Fig. S4. The multiple source dataset for discovering KG equation.** These seven datasets are solved from different initial conditions (ICs).

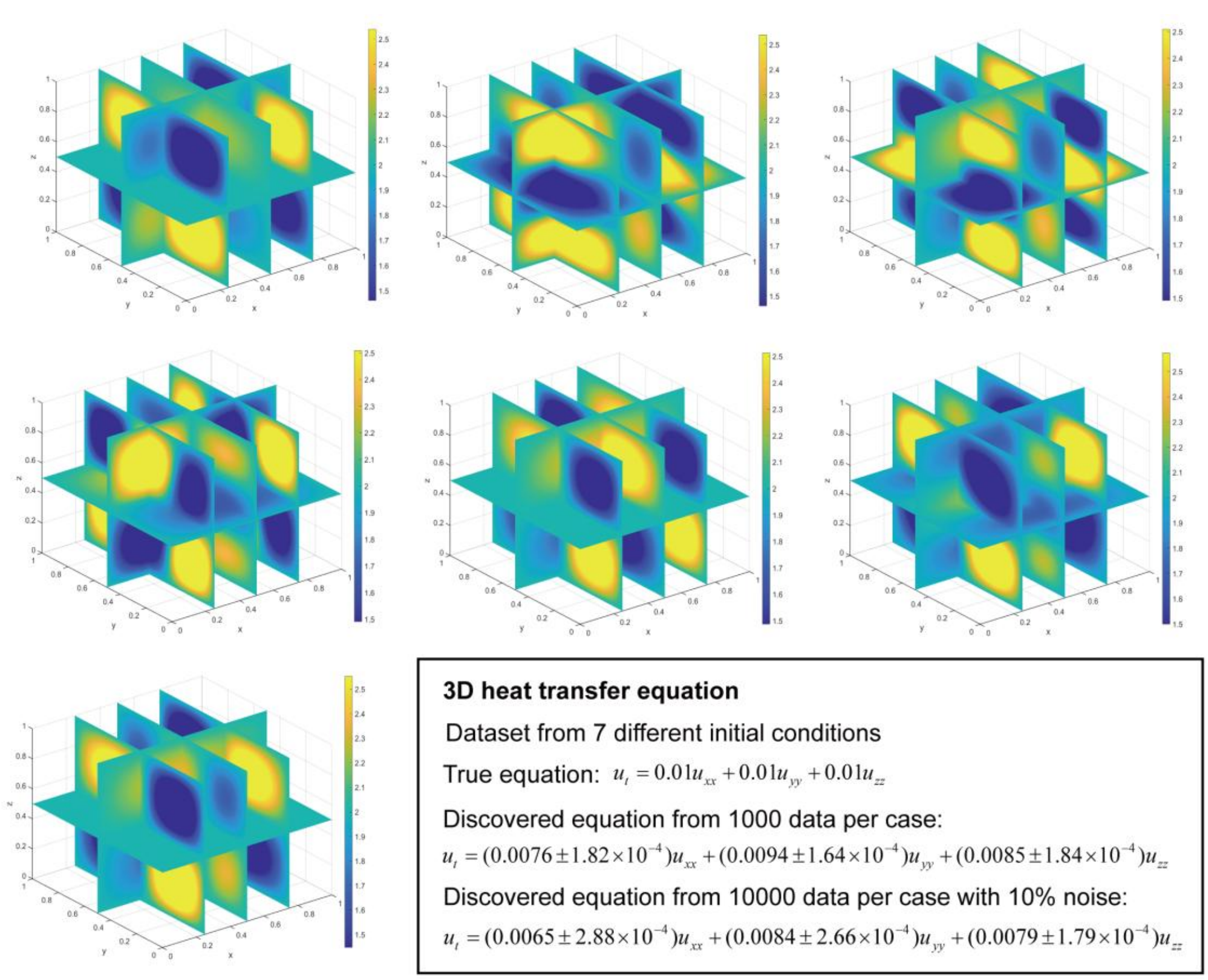

**Fig. S5 Discovery of governing equation for 3D heat transfer process from multiple source dataset.** The datasets are solved from 7 different initial conditions and the discovered equation from 1,000 data per case and 10,000 data per case is provided.

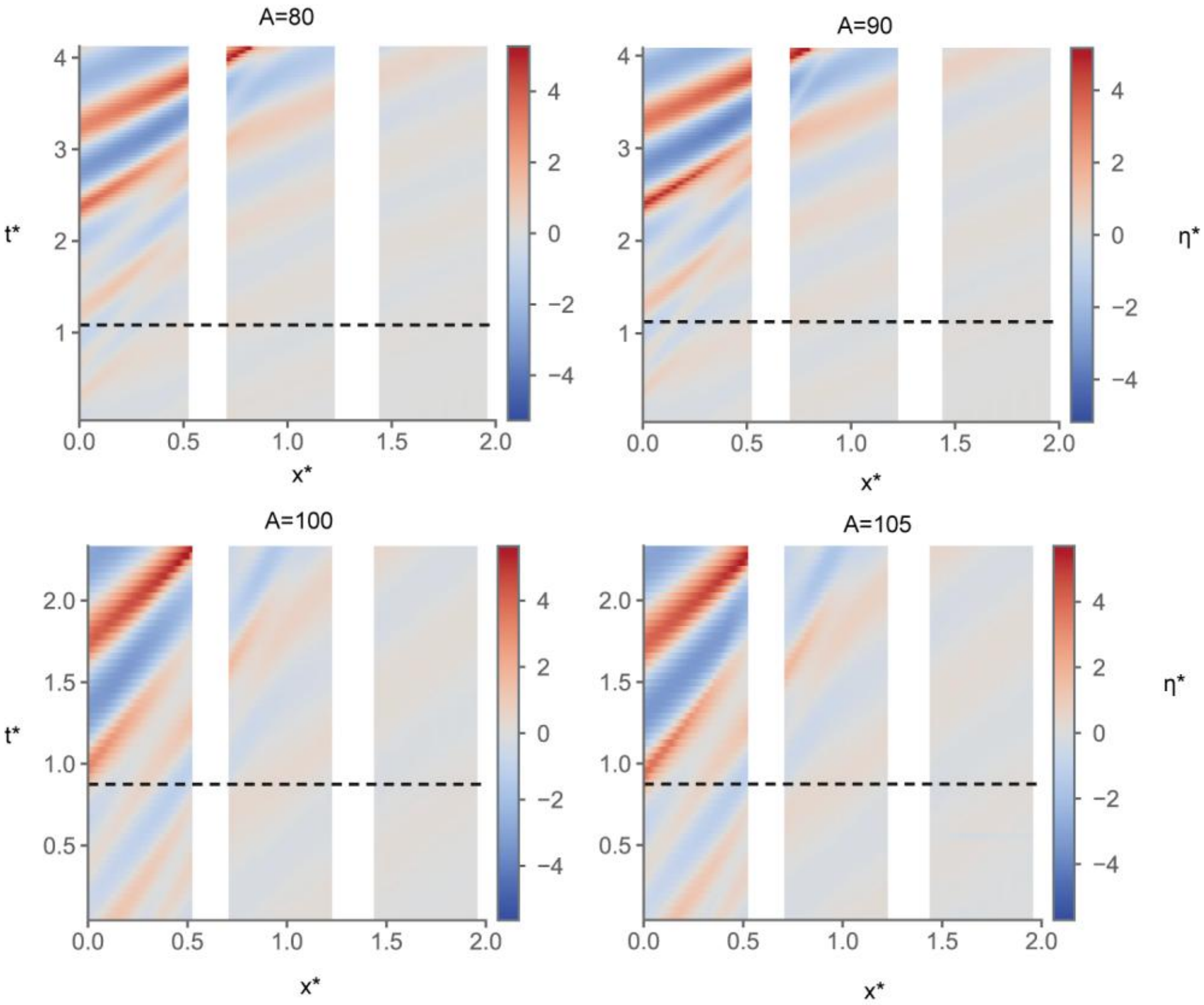

**Fig. S6 The heatmap of the dataset from real-world wave tank experiments for the discovery of wave propagation equation during pre-steepening and pre-breaking stage.** The dataset is collected from the experiments with different paddle amplitudes $A$=80, 90, 100 and 105. The black dashed line marks the transition from the pre-steepening stage to the pre-breaking stage.

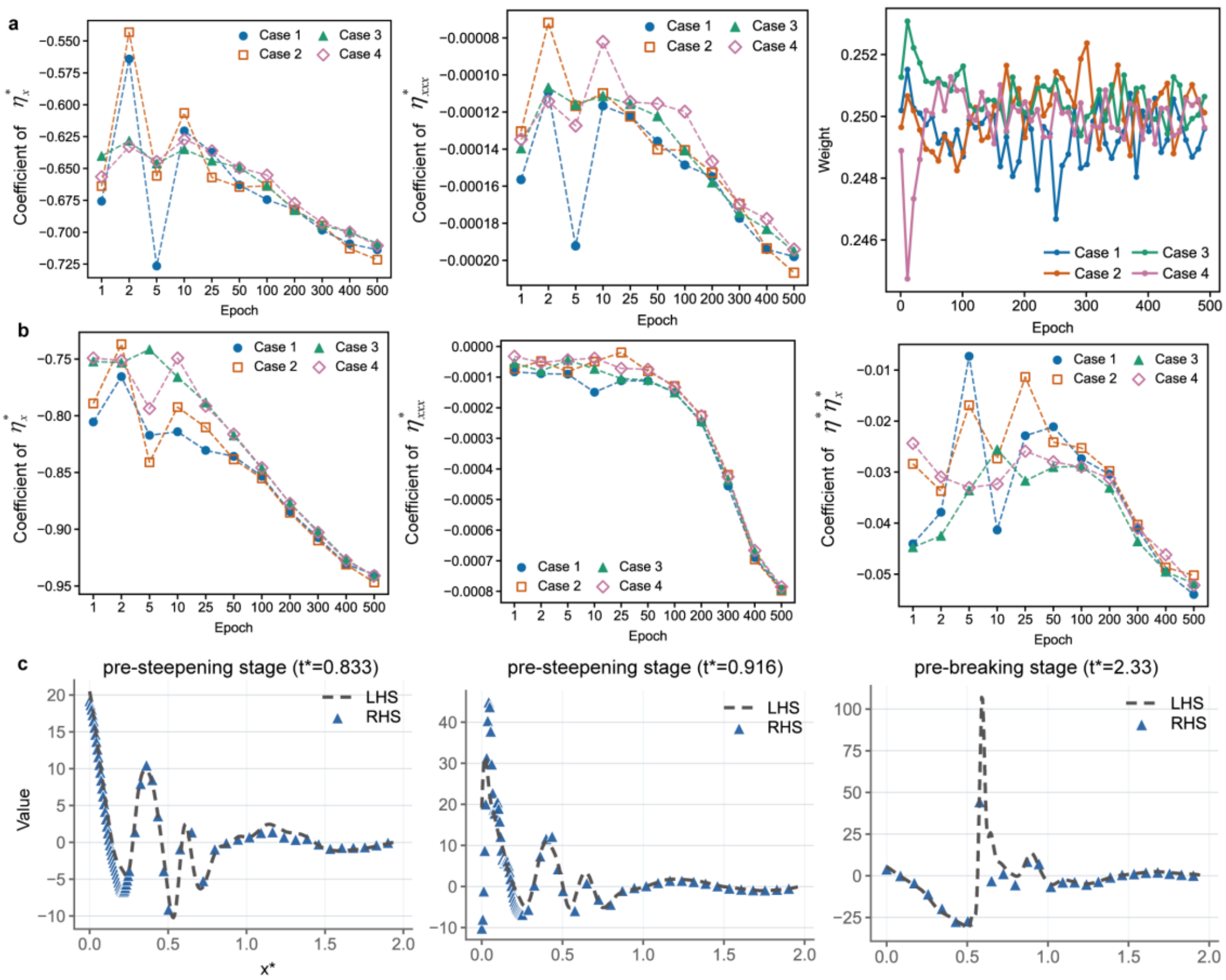

**Fig. S7. Coefficient convergence and term validation for the wave equations discovered from multi-source experimental data. (a),** Evolution of the identified coefficients and case weights during training for the pre-steepening stage. **(b),** Evolution of the identified coefficients and case weights during training for the pre-breaking stage. **(c),** Representative comparisons between the left-hand side and right-hand side of the discovered equations at selected times in the pre-steepening and pre-breaking stages.

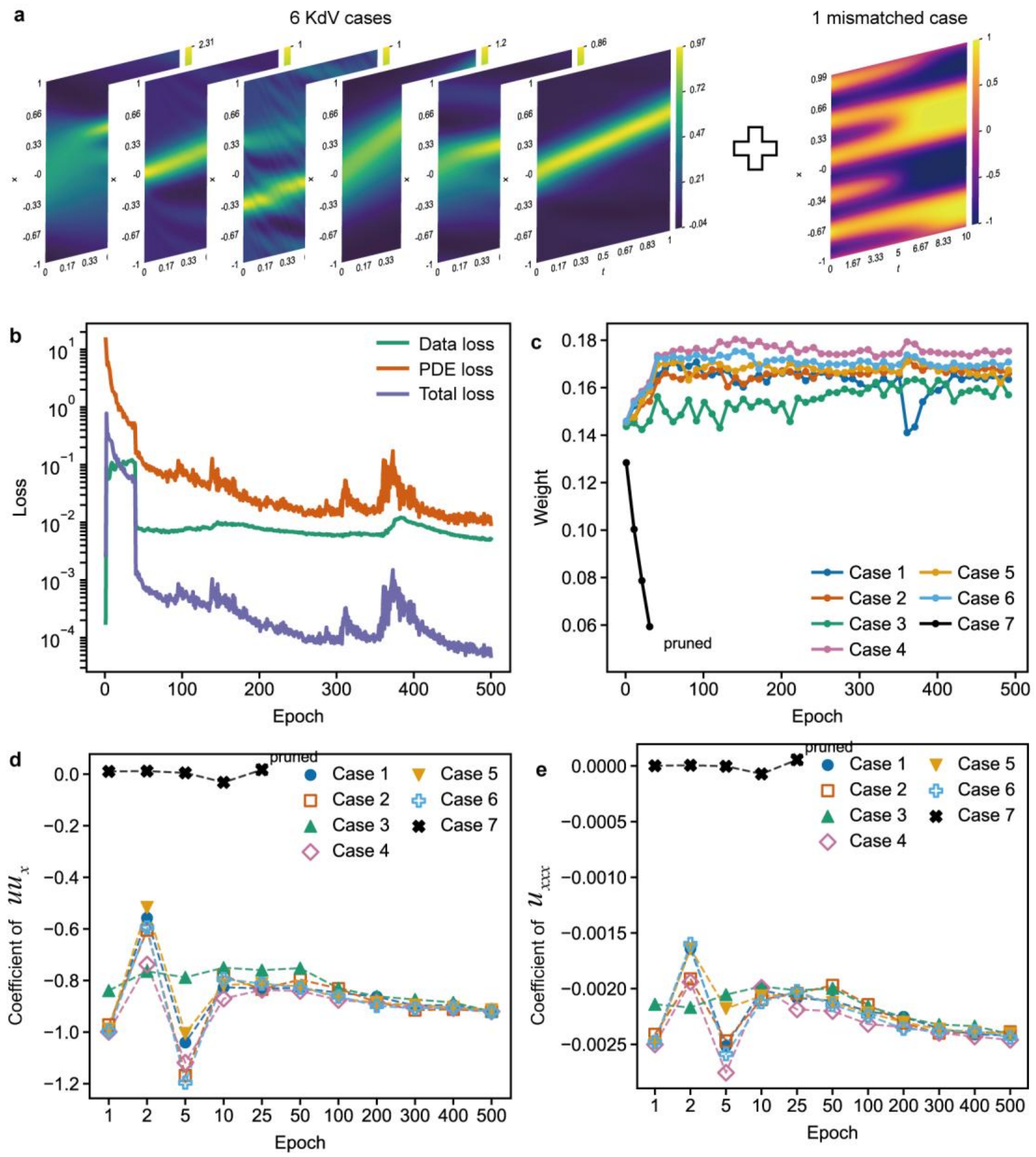

**Fig. S8. Robustness of multi-source discovery to a mismatched case. (a),** Synthetic multi-source dataset consisting of six KdV cases and one deliberately mismatched case used to test the robustness of the discovery framework, MCO-PDE. **(b),** Training curves with the fault-tolerance mechanism, showing the evolution of the data loss, PDE loss and overall loss. **(c),** Evolution of the competitive weights assigned to each case during training, showing the progressive suppression and pruning of the mismatched case. **(d,e),** Evolution of the identified coefficients for the nonlinear and dispersive terms, respectively.

**Table S1 Effect of dataset count and per-dataset data volume on multi-source discovery of the KdV equation.** Discovered equations obtained under different combinations of dataset number and data volume per dataset.

| Number of datasets | Data volume per dataset | Discovered equation |
|---|---|---|
| 1 | 50 | $\boldsymbol{u}_t = -0.640\boldsymbol{u} + 0.034\,\boldsymbol{u}_x$ |
| 1 | 100 | $\boldsymbol{u}_t = -0.424\boldsymbol{u} - 0.090\,\boldsymbol{u}_x - 0.0007\boldsymbol{u}_{xx}$ |
| 1 | 500 | $\boldsymbol{u}_t = -0.936\,\boldsymbol{u}\boldsymbol{u}_x - 0.0024\,\boldsymbol{u}_{xxx}$ |
| 2 | 50 | $\boldsymbol{u}_t = (-0.583 \pm 0.003)\,\boldsymbol{u}$ |
| 2 | 100 | $\boldsymbol{u}_t = (9.72 \times 10^{-4} \pm 3.03 \times 10^{-5})\,\boldsymbol{u}_{xxx}$ |
| 2 | 150 | $\boldsymbol{u}_t = (-0.892 \pm 0.0014)\,\boldsymbol{u}\boldsymbol{u}_x$ |
| | | $+(-0.0023 \pm 1.82 \times 10^{-5})\,\boldsymbol{u}_{xxx}$ |
| 3 | 50 | $\boldsymbol{u}_t = (-0.363 \pm 0.011)\,\boldsymbol{u}$ |
| 4 | 50 | $\boldsymbol{u}_t = (-0.811 \pm 0.017)\,\boldsymbol{u}\boldsymbol{u}_x$ |
| | | $+(-0.0021 \pm 3.89 \times 10^{-5})\,\boldsymbol{u}_{xxx}$ |
| 5 | 50 | $\boldsymbol{u}_t = (-0.160 \pm 0.002)\,\boldsymbol{u}\boldsymbol{u}_x$ |
| 6 | 50 | $\boldsymbol{u}_t = (-0.840 \pm 0.009)\,\boldsymbol{u}\boldsymbol{u}_x$ |
| | | $+(-0.0023 \pm 9.94 \times 10^{-6})\,\boldsymbol{u}_{xxx}$ |
| 7 | 50 | $\boldsymbol{u}_t = (-0.867 \pm 0.008)\,\boldsymbol{u}\boldsymbol{u}_x$ |
| | | $+(-0.0023 \pm 4.32 \times 10^{-5})\,\boldsymbol{u}_{xxx}$ |